\documentclass[10pt,journal,compsoc]{IEEEtran}

\ifCLASSOPTIONcompsoc
  \usepackage[nocompress]{cite}
\else
  \usepackage{cite}
\fi

\ifCLASSINFOpdf
\else
\fi

\usepackage{graphicx}
\usepackage{amsmath}
\usepackage{amssymb}
\usepackage{booktabs}
\usepackage{multirow}
\usepackage{diagbox}
\usepackage{bm}
\usepackage{bbm}
\usepackage{mathrsfs}

\usepackage[utf8]{inputenc} 
\usepackage[T1]{fontenc}    
\usepackage[hidelinks]{hyperref}       
\usepackage{url}            
\usepackage{booktabs}       
\usepackage{amsfonts}       
\usepackage{nicefrac}       
\usepackage{microtype}      
\usepackage{xcolor}         

\usepackage[linesnumbered,ruled]{algorithm2e}
\usepackage[capitalize]{cleveref}

\newcommand{\xiaoli}{}
\begin{document}
%
\title{Recognizing Object by Components with Human Prior Knowledge Enhances Adversarial Robustness of Deep Neural Networks}
%
%
%
%

\author{Xiao~Li,
        Ziqi~Wang,
        Bo~Zhang,
        Fuchun~Sun, ~\IEEEmembership{Fellow,~IEEE},
        and~Xiaolin~Hu,~\IEEEmembership{Senier~Member,~IEEE}

\IEEEcompsocitemizethanks{
\IEEEcompsocthanksitem{X. Li, B. Zhang, F. Sun and X. Hu are with the Department of Computer Science and Technology, Institute for Artificial Intelligence, State Key Laboratory of Intelligent Technology and Systems, BNRist, Tsinghua University, Beijing, 100084, China. Z. Wang is with the Department of Computer Science, University of Illinois Urbana-Champaign, IL, USA. $^\dag$X. Hu is the corresponding author. $^\ddag$X. Li and Z. Wang contribute equally. Email: lixiao20@mails.tsinghua.edu.cn; ziqiw9@illinois.edu; dcszb@mail.tsinghua.edu.cn; fcsun@tsinghua.edu.cn;\qquad xlhu@mail.tsinghua.edu.cn.}
}
}

%
%

%

\IEEEtitleabstractindextext{%
\begin{abstract}
Adversarial attacks can easily fool object recognition systems based on deep neural networks (DNNs). Although many defense methods have been proposed in recent years, most of them can still be adaptively evaded. One reason for the weak adversarial robustness may be that DNNs are only supervised by category labels and do not have part-based inductive bias like the recognition process of humans. Inspired by a well-known theory in cognitive psychology -- recognition-by-components, we propose a novel object recognition model ROCK (Recognizing Object by Components with human prior Knowledge). It first segments parts of objects from images, then scores part segmentation results with predefined human prior knowledge, and finally outputs prediction based on the scores. The first stage of ROCK corresponds to the process of decomposing objects into parts in human vision. The second stage corresponds to the decision process of the human brain. ROCK shows better robustness than classical recognition models across various attack settings. These results encourage researchers to rethink the rationality of currently widely-used DNN-based object recognition models and explore the potential of part-based models, once important but recently ignored, for improving robustness.
\end{abstract}

}

\maketitle

\IEEEdisplaynontitleabstractindextext

%
\IEEEpeerreviewmaketitle

\IEEEraisesectionheading{\section{Introduction}
\label{sec:intro}}


\IEEEPARstart{D}eep Neural Networks (DNNs) can be easily fooled by inputs with tiny deliberately designed perturbations. Such inputs are known as {\it adversarial examples} \cite{goodfellow14, szegedy13}.  Adversarial examples greatly hinder real-world applications of DNNs \cite{Eykholt18phy, zhu21}. The existence of adversarial examples suggests a significant difference between recognition mechanisms of DNNs and that of human, as human can hardly be fooled by such examples. To alleviate the threat of adversarial examples, a large number of defense methods have been proposed, e.g., image pre-processing \cite{liao18, ZhouL021}, loss function modification \cite{Pang18, Pang20}, and gradient regularization \cite{RossD18, YeatsCL21}. Nevertheless, attackers can still evade most of these methods methods with specially designed tricks, denoted as \emph{adaptive attacks} \cite{AthalyeC018, adaptive20}. Another idea for defense is to augment the training set by incorporating adversarial examples, which leads to Adversarial Training (AT) methods \cite{DongDP0020, goodfellow14, gene, athe, trades, AWP}. These methods are generally recognized as the most effective defense methods  under adaptive attack. However, AT has poor generalizability to unknown attacks \cite{atg}, and there is still a large gap in recognition accuracy between adversarial and benign examples for adversarially trained DNNs.

\begin{figure*}[!t]
 \centering
   \includegraphics[width=0.95\linewidth]{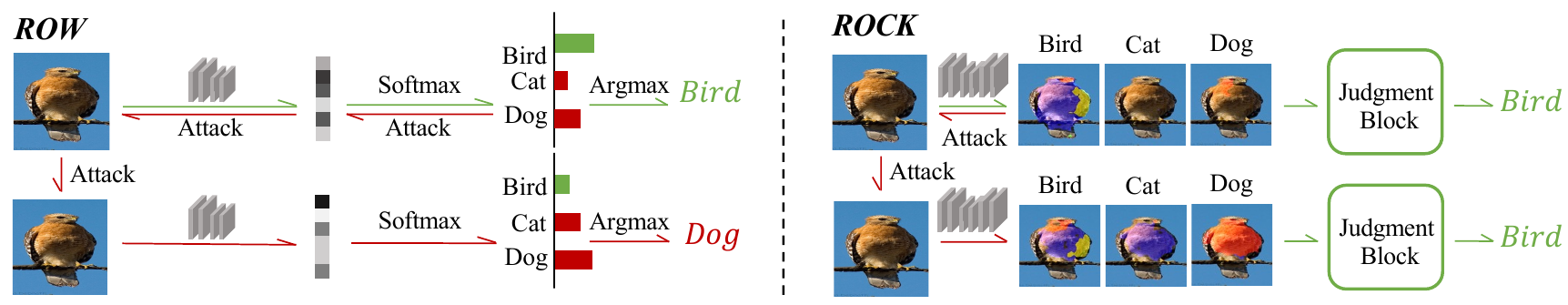}
 \caption{Comparison between \textit{ROCK} and \textit{ROW models}. Green arrows denote the recognition process of benign images, and red arrows denote the generation and recognition process of adversarial images. Given an image, no matter benign or adversarial, ROCK predicts all predefined parts represented by segmentation results, and chooses a category as the output using a judgment block.}
 \label{fig:intro}
\end{figure*}

Human visual system is much more robust against adversarial examples \cite{humanadv, imagenetp}. Imitating human vision process may help improve the robustness of object recognition models. According to recognition-by-components -- a cognitive psychological theory \cite{RBC}, human prefers to recognize objects by first dividing them into major components and then identifying the structural description and spatial relation of these components. This recognition-by-components theory has been supported by numerous psychological experiments \cite{attentioneye, oldrbc, opc}. The recognition of parts (we use ``part'' and ``component'' interchangeably in the paper) is also one of the basic skills of infants to perceive the world around them  \cite{infants}. Our work enhances the robustness of DNNs from this perspective: utilizing \textit{part-based inductive bias} and \textit{human prior knowledge}. Here, part-based inductive bias refers to recognition based on object parts. Unfortunately, the widely used recognition models nowadays based on DNNs are generally supervised by object category labels and ignore such part-based inductive bias. We refer to these models as Recognizing Object as a Whole (ROW) models in this paper. These models may inevitably recognize objects relying on some features unintelligible to human and not robust enough from the human perspective \cite{bugfeature}. This point of view is also reflected by a recent study \cite{shapetexture} in which DNNs are shown to perform recognition relying more on texture-based features while humans perform recognition relying more on shape-based features.

\begin{figure*}[!t]
  \centering
  \includegraphics[width=0.95\linewidth]{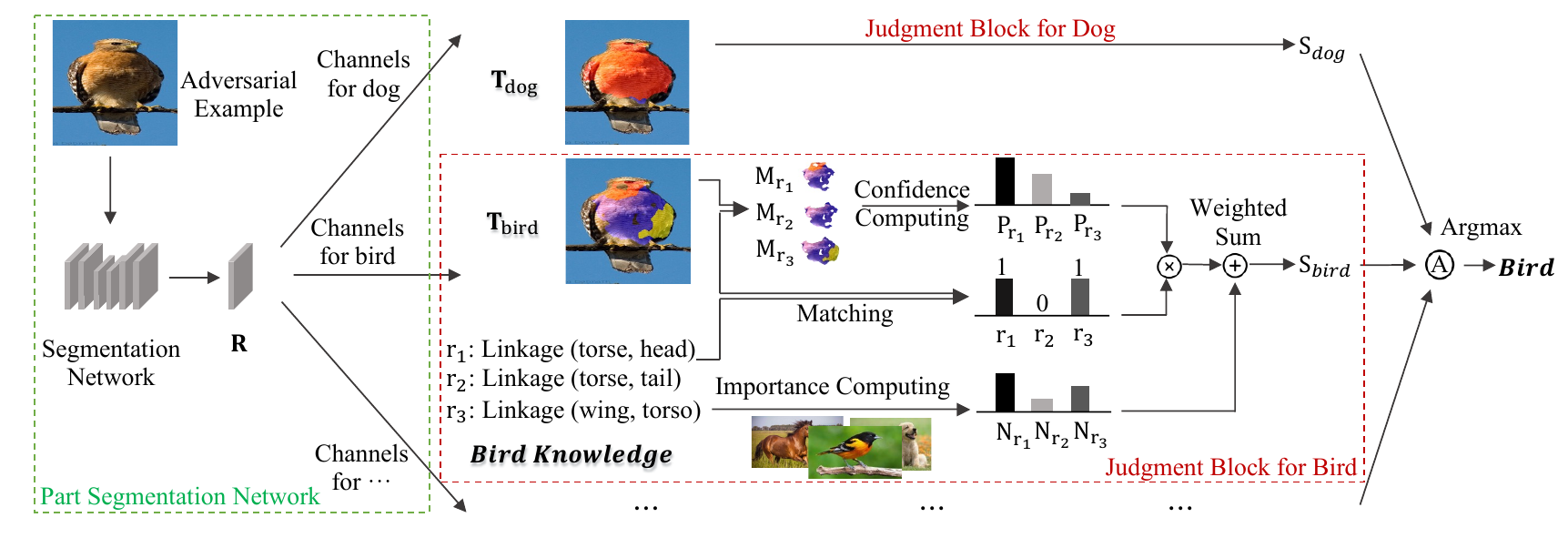}
  \caption{The pipeline of ROCK. Given an input benign or adversarial image, we first perform part segmentation for each category and then use a judgment block to evaluate each segmentation result. The category with the highest score (\textit{bird} in this image) is chosen as the output.}
  \label{fig:partseg}
\end{figure*}

Inspired by the aforementioned psychological evidence and recent observations of DNNs, we propose a novel method to enhance the robustness of DNNs by using part segmentation techniques, mimicking the human recognition process explicitly: recognizing object parts and utilizing commonsense knowledge to perform judgment. We term it ROCK, Recognizing Object by Components with human prior Knowledge.  The difference between ROCK and ROW models is shown in \cref{fig:intro}. Specifically, we supervise the recognition model by part labels to perform part segmentation. For example, when ROCK tries to recognize an object in an image as a \emph{bird}, it first segments the \emph{head}, \emph{torso}, and other parts. Then it uses a judgment block to score part segmentation results by some predefined commonsense knowledge. Finally, it recognizes the object as a \emph{bird}. With the parts segmented, it is possible to jump out of the pure data-driven feature of most machine learning methods and incorporate such human prior knowledge. \cref{fig:partseg} gives the pipeline of ROCK. The first stage of ROCK corresponds to the process of decomposing objects into parts in human vision. The second stage corresponds to the decision process of the human brain. Here we use commonsense knowledge with as little controversy as possible, such as: \emph{``if an object is a bird, its torso should be connected with its head and legs''}. We call such commonsense knowledge as topological homogeneity.


 The idea of using parts for object recognition has a long history in computer vision. Before the resurgence of DNNs, a part was often  represented by templates and other local features, and statistical models were used to recognize objects based on these representations \cite{partpro, partsl, partdis, partlearning}. Nowdays those part-based recognition models have been largely overlooked because they can not compete with DNNs in terms of accuracy. Unlike those earlier methods, ROCK maintains the advantage of their human-like recognition process while leveraging the strong representation ability of DNNs.

To show the effectiveness of ROCK, we performed various kinds of adversarial attacks. Conventional gradient-based attacks such as PGD \cite{pgd} can not attack ROCK directly as the scoring algorithm of ROCK is not differentiable. Thus, we applied adaptive PGD attacks to our model. We designed five different adaptive attacks to show the robustness of ROCK. We also evaluated ROCK under transfer-based attacks and query-based attacks. Experimental results consistently showed that ROCK was more robust than ROW models. When combined with AT, ROCK also outperformed ROW models in  similar settings. Further experiments showed that ROCK learned more shape-based features. These results suggest that mimicking human perception process is a promising direction for developing more robust object recognition models. Note that one concurrent work \cite{concurrent} also shows the robustness of part-based models.

\section{Related Work}

\subsection{Adversarial Attack and Defense}


Adversarial examples are first discovered on object recognition models \cite{szegedy13}. One typical setting is that the adversarial noise is bounded by a small norm-ball $||\cdot||_p \leq \epsilon$ so that humans can not perceive it. Several white-box adversarial attack methods in this setting have been developed \cite{goodfellow14, pgd, deepfool, mim, madry17cw}, most of which have a similar principle: crafting adversarial examples based on the input gradient \cite{benchmark}. Massive early works \cite{ob1, ob2} use obfuscated gradients to defend against such attacks but they can still be evaded through specially designed tricks \cite{AthalyeC018}. After that, the \emph{adaptive attack} has become the standard to evaluate the adversarial defense methods. In an adaptive attack, the defense method is transparent to the adversary. Therefore, the adversary can exploit this knowledge to find the weaknesses in the defense. Tram{\`{e}}r et al. \cite{adaptive20} show that quite a few recent defenses perform adaptive attacks incompletely and can still be circumvented. Except for white-box attacks, black-box attacks consider a more practical setting where the adversary does not have access to the whole model and cannot obtain the input gradient directly. Some query-based black-box attacks such as SPSA \cite{SPSA} and NES \cite{NES} try to estimate the gradient by finite difference. Among adversarial defense methods, AT is generally considered to be the most effective defense strategy. Lots of recent works \cite{trades, AWP, athe} focus on improving AT from various aspects.

Adversarial examples are also demonstrated in  semantic segmentation and object detection tasks \cite{dag, segadv}. Xie et al. \cite{dag} propose DAG to attack the detection and segmentation models iteratively. A few works \cite{ArnabMT18, character} claim that some techniques of semantic segmentation models (e.g., multiscale processing) seem to enhance the robustness of CNNs. However, they do not provide reliable experimental evidence or show the robustness of the segmentation paradigm itself.

\subsection{Part-Based Recognition}

Before the resurgence of DNNs, many object recognition methods \cite{partpro, partsl, partdis, partlearning} represented objects in terms of parts arranged in a deformable configuration. They represented a part by a template or other local features such as wavelet-like feature \cite{wavelet} and SIFT \cite{sift}. Statistical models were used to recognize objects based on part representation. For example, Burl et al. \cite{partpro} modeled objects as random constellations of parts, and the recognition was carried out by maximizing the sum of the shape log-likelihood ratio and the responses to part detectors. Crandall et al. \cite{partsl} proposed statistical models for recognition by explicitly modeling the spatial priors of parts. However, since the rise of the end-to-end models based on DNNs, those part-based recognition models have been largely overlooked. Parts have been rarely used as auxiliaries or explicit intermediate representations of general object recognition models except in some few-shot learning settings \xiaoli{\cite{compas, acn, wsod}}. Parts get attentions only in some specialized tasks, e.g., object parsing \cite{lip, Lee0W020, GMNet, bound} and 3D object understanding \cite{cgpart, partnet}. \xiaoli{Another series of work on modeling parts is capsule networks \cite{capsulesr, porvsr, onfocusr}, which use capsules to replace traditional neurons in DNNs to encode spatial relations as well as the existence of parts. However, capsule networks encode these features in vectors in an implicit way and are hard to generalize to large datasets like ImageNet \cite{imagenet}. Several recent studies \cite{capsulenor, advcapsule} show that capsule networks do not have advantages over traditional DNNs on adversarial robustness.}

\subsection{Knowledge Enhanced Machine Learning}

Most ROW models have data-driven nature. One of the modules of ROCK utilizes human prior knowledge, which makes ROCK to be a hybrid system with both data- and knowledge-driven techniques. Recently the combination of data- and knowledge-driven approaches has drawn attention. A related topic is neural-symbolic AI \cite{neuralsym}. It is believed that incorporating prior knowledge into machine learning models can make them more interpretable \cite{informedml}. To the best of our knowledge, only one work \cite{KEMLP} incorporated domain knowledge to improve robustness of DNNs on traffic sign recognition. That work uses multiple independent models, which are parallel to a DNN, to capture prior knowledge to make a decision, while ROCK uses prior knowledge after the output of a DNN to make a decision. ROCK further validates that the combination of data- and knowledge-driven approaches, even with simple human prior knowledge, has advantages over pure data-driven approaches on general object recognition.

\begin{figure*}[!t]
  \centering
  \includegraphics[width=0.95\linewidth]{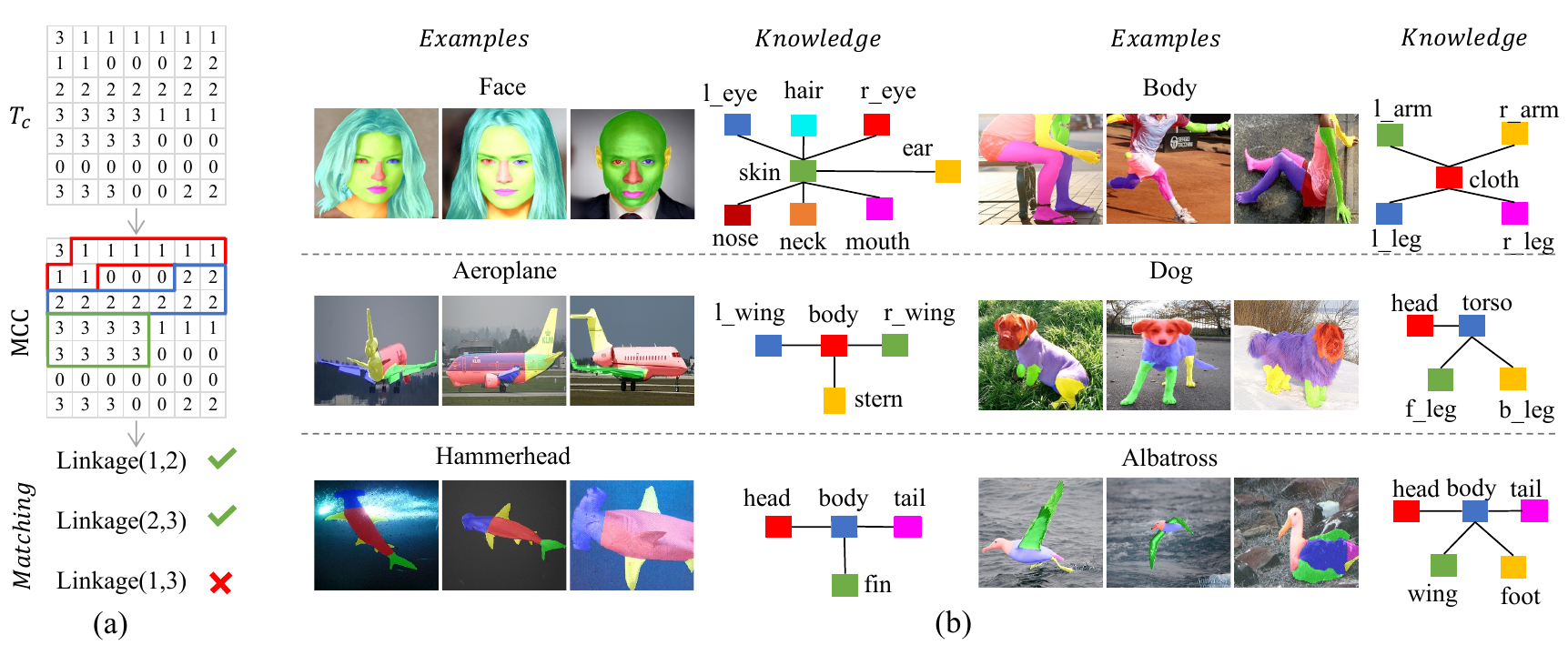}
  \caption{Illustration of knowledge matching. (a) A toy example of knowledge matching. The number in each lattice of $\mathbf{T}_c$ indicates the part label at the corresponding location. First, the MCC is calculated for each part, and then these MCCs are judged by linkage rules. (b) Examples from Face-Body (upper), Pascal-Part-C (middle), and PartImageNet-C (lower). To the right of the images, colored blocks denote annotated object parts and lines between them denote predefined linkage rules abstracted from commonsense knowledge.}
  \label{fig:examplemerge}
\end{figure*}

\section{ROCK}

The pipeline of ROCK is illustrated in \cref{fig:partseg}. ROCK recognizes objects in two stages. It first uses a segmentation network to segment the object into parts of different object categories. Then it utilizes a judgment block to score each segmentation result and perform recognition over the object.

\subsection{Part Segmentation Network}
\label{sec:psn}
Suppose that we have an image dataset $\{\mathbf{I}\}$ of $C$ categories with object labels $\{\mathbf{y_c}\}$. For each object category $c$, we specify a set of part categories. These part category sets are disjoint. In this work, we assume all objects to have several parts (see \nameref{sec:diss}). The part labels of all object categories are denoted by $\{\mathbf{y_p}\}$. A ROW method usually trains a DNN by minimizing a softmax cross-entropy loss based on the output of the DNN and $\{\mathbf{y_c}\}$. ROCK does not use object labels in this way. Instead, we train a part segmentation network $F$ parameterized by $\theta_{F}$, where $F$ can be an arbitrary fully convolutional network (FCN \cite{long2015fully}). $F$ generates part response maps $\mathbf{R}={\rm softmax}(F\left(\mathbf{I} ; \theta_{F}\right)) \in [0,1]^{(K + 1) \times H \times W}$, where $K$ denotes the number of parts (channels) and $H \times W$ denotes the resolution. The extra one channel indicates the background. During the training procedure, the FCN is trained in the same way as in usual segmentation tasks supervised by $\{\mathbf{y_p}\}$. The key of ROCK is to obtain recognition outputs from the segmentation results in the inference stage.

During the inference stage, we obtain the part response maps $\mathbf{R}$ by the segmentation network. Let $\mathbf{B} \in [0,1]^{H \times W}$ denote the background channel. We then split $\mathbf{R}$ with $K + 1$ channels into groups and channels in the same group correspond to parts belonging to the same object category. We denote channels of the same group as $\mathbf{R}_{c} \in [0,1]^{K_c \times H \times W}$ and all channels' indices of $\mathbf{R}_{c}$ as $\Omega_c = \{i_1,\cdots,i_{K_c}\}$, where $K_c$ denotes the number of part categories in object category $c$ and $\cup_c \Omega_c = \{1, \cdots, K\}$. We name $\mathbf{R}_{c}$ as part response for $c$ and $\Omega_c$ as part set of $c$. After generating the part response for all categories, each $\mathbf{R}_{c}$ is evaluated by a judgment block.

\subsection{Judgment Block}
The judgment block is a module that uses human prior knowledge to score segmentation results. We use the concept \textit{linkage} to build topology-based commonsense knowledge rules. The final category prediction of ROCK is the category that best matches these rules, e.g., if we detect a \textit{bird head} and a \textit{bird torso}, this object has a higher probability of being predicted as a \textit{bird} if these two parts are \textit{linked}. Some rules are listed in \cref{fig:examplemerge}. More can be found in Appendix \ref{sec:apd_dataset}. The concept \textit{linkage} can be applied if and only if corresponding parts are segmented out. So we first detect parts from the part segmentation results. On each $\mathbf{R}_{c}$, a part label $\mathbf{T}_c(i,j)$ at each location is predicted:
\begin{equation}
\mathbf{T}_c(i,j)=\left\{
\begin{aligned}
0,\quad{\rm if} \quad \mathbf{B}(i,j)\geq\sum\nolimits_{k \in \Omega_c} \mathbf{R}_c(k,i,j)\\
\arg\max\nolimits_{k \in \Omega_c}\mathbf{R}_c(k,i,j), \quad {\rm otherwise.}
\end{aligned}
\right.
\label{eq:fb}
\end{equation}
 According to this definition, $\mathbf{T}_c(i,j) > 0$ indicates that a part whose label belongs to category $c$ is present at location $(i,j)$. We then calculate the Maximal Connected Component (MCC) for each predicted part label $p$ in $\mathbf{T}_c$ (\cref{fig:examplemerge}(a) shows three MCCs for three part labels), denoted by a mask $\mathbf{M}_p \in \{0,1\}^{H \times W}$. For each part label, we keep the MCC as the real part area and remove other connected components considering that part segmentation results inevitably have false positives.

Let $\mathbf{M}_{p_1}$ and $\mathbf{M}_{p_2}$ denote the masks of the MCCs of two parts that are in the same $\mathbf{T}_c$. These two parts are considered to be \textit{linked} if and only if $\mathbf{M}_{p_1} \cup \mathbf{M}_{p_2}$ is a connected component. As \cref{fig:examplemerge}(b) shows, each category has a predefined set of linkage rules abstracted from commonsense knowledge. If two detected parts in $\mathbf{T}_c$ are linked, and this rule appears in the predefined set of rules of category $c$, we say that this rule matches category $c$. Let $\mathbf{M}_{r}$ denote the sum of two part masks involved in a linkage rule $r$, $\mathbf{V}_c(i,j) = \max_{k \in \Omega_c}\mathbf{R}_c(k,i,j)$ denotes the maximal part response where the maximum operation is applied on channels, then the matching confidence of rule $r$ is computed as:
\begin{equation}
\begin{aligned}
  P_{r}^{c} &= \sum_{i=1,j=1}^{H,W} \mathbf{V}_{c}(i,j) \cdot \mathbf{M}_{r}(i,j). \\
\end{aligned}
\label{eq:s2}
\end{equation}


Clearly, if more detected linkage rules in an image match a category and the matched linkage rules of the category have greater matching confidence, that category is more likely to be the real category to which the image belongs. So an intuitive scoring strategy is to compute the final score for category $c$ as the sum of the matching confidence of detected linkage rules in $\mathbf{T}_c$: $\sum_{r \in \mathcal{L}_c} \mathbbm{1}_{[{\rm match}(r, c)]} \cdot P_r^c$, where $\mathcal{L}_c$ denotes the predefined linkage rule set of category $c$, $\mathbbm{1}_{[{\rm match}(r, c)]}$ is one if the linkage rule $r$ matches category $c$, and zero otherwise. However, this strategy neglects that different linkage rules are of different importance. For example, a \textit{head}-\textit{torso} \textit{linkage} may be more important than a \textit{torso}-\textit{wing} \textit{linkage} to help human recognize \textit{bird}, as sometimes a \textit{wing} is not that easily noticed. Besides, some parts like \textit{cat leg} are often obscured in real images. So we use the linkage distribution in the training set to estimate the importance of different linkage rules, and the final score of $c$ is computed by:
\begin{equation}
\begin{aligned}
  S_c &= \frac{\sum_{r \in \mathcal{L}_c} N_r \cdot \mathbbm{1}_{[{\rm match}(r, c)]} \cdot P_r^c} {\sum_{r \in \mathcal{L}_c} N_r} \\
\end{aligned}
\label{eq:s3}
\end{equation}
where $N_r$ denotes the number of occurrence of the linkage rule $r$ in the training set. Here, we assign each linkage rule $r$ of category $c$ a weight $\frac{N_r}{\sum_{r \in \mathcal{L}_c} N_r}$. If the linkage rule $r$ rarely appears in the training set, its weight contributes little to $S_c$. Finally, the final prediction for image $\mathbf{I}$ is $c^{\star} = \arg\max_{c \in \{1,\cdots,C\}} S_c$.

\begin{table*}[!t]
  \centering
  \caption{Classification accuracies (\%)  under transfer-based attacks. Adversarial examples were generated on \textit{source} models and evaluated on \textit{target} models.}
  \small
   \resizebox{\linewidth}{!}{
   \setlength{\tabcolsep}{1.5pt}
   {
    \begin{tabular}{c|c||cccc||cccc||cccc}
    \toprule
    \multirow{2}*{\textbf{Source}}   & \multirow{2}*{\textbf{Target}}  & \multicolumn{4}{c||}{\textbf{Face-Body}} & \multicolumn{4}{c||}{\textbf{Pascal-Part-C}} & \multicolumn{4}{c}{\textbf{PartImageNet-C}} \\
    \cmidrule{3-14}
    & &\textbf{PGD-40} & \textbf{FGSM}& \textbf{MIM-20}& \textbf{PGD-400} & \textbf{PGD-40} & \textbf{FGSM}& \textbf{MIM-20}& \textbf{PGD-400} & \textbf{PGD-40} & \textbf{FGSM}& \textbf{MIM-20}& \textbf{PGD-400} \\
    \midrule
    \multirow{3}*{ROW-50}     & ROW$\rm{_X}$-101      & $69.9$   & $92.0$  & $69.2$  & $70.9$ & $29.9$   & $35.8$  & $23.0$  & $31.5$ & $9.3$   & $13.9$  & $5.4$  & $10.1$  \\ 
                                  &  ROW$\rm{_M}$-50    & $30.4$   & $85.9$  & $45.9$  & $29.7$  & $23.4$   & $34.0$  & $20.0$  & $25.9$ & $12.0$   & $16.9$  & $7.6$  & $13.4$ \\
                                   &  ROCK      & $\bm{98.5}$   & $\bm{99.5}$  & $\bm{99.1}$  & $\bm{98.5}$ & $\bm{50.0}$   & $\bm{52.2}$  & $\bm{46.8}$  & $\bm{50.0}$   & $\bm{33.9}$   & $\bm{25.4}$  & $\bm{26.1}$  & $\bm{33.4}$\\
    \cmidrule{1-14}
     \multirow{3}*{ROW$\rm{_X}$-101}     & ROW-50      & $71.8$   & $87.0$  & $61.1$  & $74.8$ & $32.6$   & $39.0$  & $27.4$  & $35.7$  & $12.4$   & $16.5$  & $10.4$  & $12.6$  \\
                                  & ROW$\rm{_M}$-50      & $61.6$   & $92.4$  & $71.0$  & $64.3$ & $22.5$   & $34.9$  & $20.0$  & $23.9$ & $18.6$   & $22.4$  & $14.9$  & $19.6$\\
                                  & ROCK      &  $\bm{99.2}$   & $\bm{99.6}$  & $\bm{98.9}$  & $\bm{99.0}$ &$\bm{46.1}$   & $\bm{52.1}$  & $\bm{41.0}$  & $\bm{47.0}$ & $\bm{31.8}$   & $\bm{25.3}$  & $\bm{26.0}$  & $\bm{31.6}$\\
    \cmidrule{1-14}

     \multirow{3}*{ROW$\rm{_M}$-50}       & ROW-50      & $40.3$   & $84.3$  & $44.3$  & $40.6$ & $13.6$   & $24.7$  & $13.5$  & $13.9$ & $13.4$   & $14.3$  & $8.3$  & $14.5$ \\
                                  & ROW$\rm{_X}$-101      & $65.4$   & $92.1$  & $69.3$  & $65.8$ & $11.1$   & $23.2$  & $8.1$  & $10.1$ & $17.0$   & $17.2$  & $10.2$  & $17.9$  \\
                                  & ROCK      & $\bm{98.3}$  & $\bm{99.6}$  & $\bm{98.7}$ & $\bm{98.6}$ & $\bm{29.5}$   & $\bm{41.7}$  & $\bm{27.1}$  & $\bm{29.8}$ & $\bm{21.5}$ & $\bm{18.2}$  & $\bm{14.2}$  & $\bm{21.6}$\\

    \bottomrule
    \end{tabular}
    }
    }

  \label{tab:transfer}%
\end{table*}%

\section{Experiments}
\label{sec:expr}

\subsection{Datasets}
Pascal-Part \cite{parttoc} and PartImageNet \cite{partimagenet} are the only two existing datasets that offer part labels on a general set of categories to our best knowledge, but they were designed for object detection and part segmentation, respectively. We adjusted them for object recognition and call them Pascal-Part-C and PartImageNet-C, where ``C'' stands for classification. In addition, we collected a new Face-Body object recognition dataset from other datasets. Here we briefly describe these datasets and more details can be found in Appendix \ref{sec:apd_dataset}. We will release Pascal-Part-C and PartImageNet-C and the code of ROCK to encourage follow-up works after the paper is revised\footnote{Face-Body is restricted by the license, but we will release the processing code to build it from the original datasets. }.

The Face-Body dataset was created from two open datasets: CelebAMask-HQ \cite{Lee0W020} and LIP \cite{lip}. The former has faces with facial part segmentation labels such as \emph{mouth} and \emph{ear}. The latter has body part segmentation labels such as \emph{arm} and \emph{leg}. We used CelebAMask-HQ as one category \emph{face}. As images of persons contain faces, we only used body parts of LIP images as the category \emph{body}. Finally, we got 11,767 body images and 11,767 face images. See \cref{fig:examplemerge}(b) (upper) for examples.

The Pascal-Part-C dataset was built from the Pascal-Part dataset \cite{pascal2010}, which was originally used for the object detection task. We used images and labels from 10 categories: \emph{person}, \emph{cat}, \emph{car}, \emph{bird}, \emph{dog}, \emph{aeroplane}, \emph{bus}, \emph{horse}, \emph{train}, and \emph{sheep}. Pascal-Part contains multiple categories per image, so we extracted objects with little overlap according to object bounding boxes. A total of 8,609 images with 37 different kinds of part segmentation labels were kept.  See \cref{fig:examplemerge}(b) (middle) for examples. 


The PartImageNet-C dataset was built from the PartImageNet dataset \cite{partimagenet}, which was originally used for part segmentation. PartImageNet consists of 24,095 images of 158 object categories. We removed categories with images fewer than 100. We found that the labels of certain categories are not suitable for part segmentation, e.g., \emph{car} is annotated with \emph{side mirror}. We removed these categories. A total of 20,509 images from 125 categories were kept. See \cref{fig:examplemerge}(b) (lower) for examples.

\subsection{Experimental Settings}

In all experiments, we used ResNet-50 \cite{resnet} and ResNeXt-101 \cite{ResNeXt} as two baselines, denoted by ROW-50 and ROW$\rm{_X}$-101 respectively. For comparison, the segmentation network we used in ROCK was a slightly modified ResNet-50 (denoted by ROW$\rm{_M}$-50). The main difference between ROW$\rm{_M}$-50 and ROW-50 is that ROW$\rm{_M}$-50 changed the first 3 $\times$ 3 convolutions in each B3 and B4 to be two dilated convolutions with the dilation of 2 and 4 respectively, to obtain the segmentation output with the resolution of 28 $\times$ 28. And to reduce the extra computational cost introduced by larger feature maps in B3 and B4, ROW$\rm{_M}$-50 reduced the width from 256 and 512 to 192 and 384, respectively, and used depth-wise separable convolutions \cite{deepwise} to replace dense convolutions in B3 and B4. We also compared ROCK with ROW$\rm{_M}$-50. 

We performed experiments on the three datasets mentioned above, considering that existing popular datasets for image classification task are missing part labels. The input resolution of 224 $\times$ 224 was used, as that on the ImageNet dataset \cite{imagenet}. Note that ROCK does not rely on precise segmentation results to work, so during training, we downsampled the part labels to 28 $\times$ 28 to lower down computational cost. Moreover, we only matched top-10 categories ranked by the foreground size of $\mathbf{T}_c$ if the number of categories was large. As a result, the extra knowledge matching cost was quite low (the time complexity is $\mathcal{O}(HW)$). \xiaoli{With these designs, ROCK and ROW$\rm{_M}$-50 had comparable theoretical computational cost (\cref{tab:flops}). The actual training time is basically consistent with the theoretical Flops.}

Unless otherwise specified, all models were optimized using stochastic gradient descent with an initial learning rate of 0.1 and a momentum of 0.9. Data augmentation included random flipping and cropping. \xiaoli{The normalization of the inputs used a mean of 0.5 and a standard deviation of 0.5.} All models were trained from scratch with random initialization for 200 epochs by using 4 RTX 3090 GPUs. Following \cite{deeplabv2}, ROCK used polynomial decay with a power of 0.9, while ROW models used multi-step decay that scales the learning rate by 0.1 after the 150th and 170th epochs. Focal loss \cite{focal}, a variant of softmax cross-entropy loss, was used to accelerate the convergence of the segmentation network.

\begin{table*}[!t]
  \centering
  \caption{ROCK's accuracies (\%) under various adaptive attacks. All attacks iterated 40 steps using PGD with step size $\epsilon / 8$. We highlight the \textbf{lowest} accuracies of ROCK under the five adaptive attacks.}
  \small
   \setlength{\tabcolsep}{2.5pt} {
    \begin{tabular}{c||c|cc||c|cc||c|cc}
    \toprule
    \multirow{2}*{\textbf{Method}}   & \multicolumn{3}{c||}{\textbf{Face-Body}} & \multicolumn{3}{c||}{\textbf{Pascal-Part-C}} & \multicolumn{3}{c}{\textbf{PartImageNet-C}}\\
    \cmidrule{2-10}
       & \textbf{Benign} & \textbf{$\bm{\epsilon =1}$} & \textbf{$\bm{\epsilon =2}$} & \textbf{Benign} & \textbf{$\bm{\epsilon =1}$} & \textbf{$\bm{\epsilon =2}$} & \textbf{Benign} & \textbf{$\bm{\epsilon =0.5}$} & \textbf{$\bm{\epsilon =1}$}\\
    \midrule
    \textbf{Random} & \multirow{5}*{$100$}& $89.6$& $50.1$ & \multirow{5}*{$80.7$}& $29.0$& $\bm{8.6}$ & \multirow{5}*{$66.7$}& $35.8$& $13.3$ \\
    \textbf{Targeted} & & $\bm{88.1}$ &$\bm{45.2}$ & & $36.5$ &$8.8$ & & $40.5$ & $17.8$ \\
    \textbf{Untargeted} & &$89.3$ &$51.8$  & & $35.5$ &$9.4$ & & $40.0$ &$16.7$ \\
    \textbf{Background} & &$94.8$ &$68.7$ & &  $33.1$ &$10.6$ & &  $36.7$ &$14.1$\\
   \textbf{Importance} & & $96.8$ &$73.0$ & & $\bm{28.2}$ & $8.7$ & & $\bm{31.3}$ &$\bm{10.7}$ \\
    \bottomrule
    \end{tabular}
    }
\label{tab:AttackROCK}%
\end{table*}%

\begin{table*}[!t]
  \centering
  \caption{Classification accuracies (\%) of ROCK and ROW models under the white-box attacks and AutoAttack. The lowest accuracies of ROCK under five adaptive attacks are reported here.}
  \small
  \setlength{\tabcolsep}{1pt}
  {
    \begin{tabular}{c||c|cc|cc||c|cc|cc||c|cc|cc}
    \toprule
     \multirow{3}*{\textbf{Target}} & \multicolumn{5}{c||}{\textbf{Face-Body}}  & \multicolumn{5}{c||}{\textbf{Pascal-Part-C}} & \multicolumn{5}{c}{\textbf{PartImageNet-C}} \\
     \cmidrule{2-16}
                                     & \textbf{Benign} & \multicolumn{2}{c|}{\textbf{PGD-40}} & \multicolumn{2}{c||}{\textbf{AutoAttack}} & \textbf{Benign} & \multicolumn{2}{c|}{\textbf{PGD-40}} & \multicolumn{2}{c||}{\textbf{AutoAttack}} & \textbf{Benign} & \multicolumn{2}{c|}{\textbf{PGD-40}} & \multicolumn{2}{c}{\textbf{AutoAttack}} \\
                                     & \textbf{Acc} & \textbf{$\bm{\epsilon =1}$} & \textbf{$\bm{\epsilon =2}$} & \textbf{$\bm{\epsilon =1}$} & \textbf{$\bm{\epsilon =2}$} & \textbf{Acc} & \textbf{$\bm{\epsilon = 1}$} & \textbf{$\bm{\epsilon =2}$} & \textbf{$\bm{\epsilon =1}$} & \textbf{$\bm{\epsilon =2}$} & \textbf{Acc} & \textbf{$\bm{\epsilon = 0.5}$} & \textbf{$\bm{\epsilon =1}$} & \textbf{$\bm{\epsilon =0.5}$} & \textbf{$\bm{\epsilon =1}$} \\
    \midrule
    ROW-50  & $99.9$ & $74.4$ & $14.3$ & $72.3$ & $12.0$ & $80.0$ & $3.5$ & $0.0$ & $2.9$ & $0.0$ & $62.1$ & $20.2$ & $3.9$ & $19.3$ & $3.5$\\
    ROW$\rm{_M}$-50  & $99.9$ & $84.7$ & $22.4$ & $83.5$ & $18.7$ & $80.0$ & $5.4$ & $0.0$ & $3.9$ & $0.0$  & $66.3$ & $23.1$ & $5.4$ & $22.2$ & $4.8$\\
    ROW$\rm{_X}$-101  & $99.9$ & $85.3$ & $25.6$ & $83.9$ & $18.7$ & $\bm{81.7}$ & $2.7$ & $0.0$ & $2.7$ & $0.0$  & $66.2$ & $20.5$ & $3.2$ & $19.3$ & $2.7$\\
    \midrule
    ROCK  & $100$ & $\bm{88.1}$ & $\bm{45.2}$ & $\bm{87.4}$ & $\bm{43.3}$ & $80.7$ & $\bm{28.2}$ & $\bm{8.6}$ & $\bm{23.3}$ & $\bm{5.7}$  & $\bm{66.7}$ & $\bm{31.3}$ & $\bm{10.7}$ & $\bm{28.7}$ & $\bm{9.2}$\\

    \bottomrule
    \end{tabular}
    }
  \label{tab:facebase}
\end{table*}

\subsection{Adversarial Robustness}
Following \cite{KEMLP, athe, fd}, all attacks were considered under the commonly used norm-ball $\lvert\lvert \mathbf{x} - \mathbf{x}_{\rm{adv}} \rvert\rvert_\infty \leq \epsilon/255$, which bounded the maximal difference for each pixel of an image $\mathbf{x}$. The value of $\epsilon$ was relative to the pixel intensity scale of 255. All pixels were scaled to $[0,1]$ before being attacked. 




\subsubsection{Transfer-Based Attacks}
Adversarial examples have transferability \cite{transfer} across different models, i.e., an adversarial example that is crafted for fooling one object recognition model may also fool other models. We first performed experiments to show the robustness of ROCK under transfer-based attacks based on ROW models. Here adversarial examples were generated using  MIM\cite{mim}, FGSM \cite{goodfellow14}, and PGD \cite{pgd}, and all attacks used the maximal intensity $\epsilon = 8$. \xiaoli{Note that MIM and PGD are iterative methods and the perturbation step size 1 was used.} In what follows, the number suffixes denote iteration steps. \cref{tab:transfer} shows the accuracies under transfer-based attacks on three datasets. ROCK was consistently more robust than ROW models against various transfer-based attacking methods.

\begin{table*}[!t]
\caption{\xiaoli{Classification accuracies (\%) under the query-based black-box attack. We set $\sigma$ used in SPSA and NES to be 0.001 and the number of random samples in each iteration step to be 128. Both SPSA and NES attacks iterated 40 steps. Performing them needed to query a model 5,120 times in total. RayS used 5,000 search steps.}}
\small
\centering
\setlength{\tabcolsep}{3.5pt} {
\begin{tabular}{c|cc|cc|cc|cc|cc|cc}
\toprule
\multirow{3}{*}{\textbf{Target}}  &           \multicolumn{6}{c|}{\textbf{Pascal-Part-C}}                &                       \multicolumn{6}{c}{\textbf{PartImageNet-C}}                     \\
\cmidrule{2-13}
                                  &           \multicolumn{2}{c|}{\textbf{SPSA}}                 &           \multicolumn{2}{c|}{\textbf{NES}} &     \multicolumn{2}{c|}{\xiaoli{\textbf{RayS}}}   &       \multicolumn{2}{c|}{\textbf{SPSA}}                 &           \multicolumn{2}{c|}{\textbf{NES}}       & \multicolumn{2}{c}{\xiaoli{\textbf{RayS}}}              \\
                              & \multicolumn{1}{c}{$\bm{\epsilon =4} $} & \multicolumn{1}{c|}{$\bm{\epsilon =8}$} & \multicolumn{1}{c}{$\bm{\epsilon =4}$} & \multicolumn{1}{c|}{$\bm{\epsilon =8}$}  & \multicolumn{1}{c}{$\bm{\epsilon =2} $} & \multicolumn{1}{c|}{$\bm{\epsilon =4}$}  & \multicolumn{1}{c}{$\bm{\epsilon =4} $} & \multicolumn{1}{c|}{$\bm{\epsilon =8} $} & \multicolumn{1}{c}{$\bm{\epsilon =4} $} & \multicolumn{1}{c|}{$\bm{\epsilon =8} $} & \multicolumn{1}{c}{$\bm{\epsilon =2} $} & \multicolumn{1}{c}{$\bm{\epsilon =4}$} \\
\midrule
ROW-50                   &     $33.5$                        &          $10.7$                      &                               $33.8$         &     $10.7$   & $32.1$& $9.8$&     $17.4$                     &            $4.7$                      &                               $17.4$     &     $5.4$  & $17.6$&$9.0$\\
ROW$\rm{_M}$-50          &      $34.7$                       &          $10.8$                      &                                $35.0$        &      $11.3$   &$36.3$& $12.2$&     $17.3$                     &            $5.7$                      &                               $17.9$     &     $5.4$  & $24.5$& $5.3$\\
ROW$\rm{_X}$-101         &      $22.9$                       &          $9.8$                      &                                $22.9$         &     $9.4$   & $29.9$ & $11.1$&     $19.3$                     &            $5.2$                      &                               $18.5$     &     $4.9$   &$19.5$&$5.3$\\
\midrule
ROCK                    &      $\bm{60.5}$                  &          $\bm{52.3}$                 &                                $\bm{58.0}$   &     $\bm{51.3}$  & $\bm{46.1}$& $\bm{22.5}$ &     $\bm{35.9}$                     &            $\bm{24.6}$                      &                               $\bm{37.1}$     &     $\bm{27.0}$   & $\bm{38.0}$& $\bm{16.1}$\\

\bottomrule
\end{tabular}
}
\label{tab:query}
\end{table*}

\subsubsection{White-Box Adaptive Attacks}
Adaptive attack \cite{AthalyeC018, adaptive20} must be performed before claiming real adversarial robustness. If the defense model is differentiable, it is easy to perform adaptive attack. Nevertheless, ROCK has a non-differentiable knowledge matching operation ($\mathbbm{1}_{[{\rm match}(r, c)]}$). Thus, we designed five adaptive attack methods that depend on differentiable parts of ROCK. 

\setlength{\algomargin}{0.7em}
\SetNlSkip{0.4em}
\begin{algorithm}[h]
    \SetKwInOut{Return}{Return}
    \SetKwInOut{Input}{Input}
    \SetKwInOut{Output}{Output}
        \caption{Modified DAG}
        \label{algo:algo1}
    \Input{
        image $\mathbf{x}$ with resolution $H \times W$;\\
        \ segmentation model $F(\cdot,\cdot) \in \mathbb{R}^{K+1}$; \\

        \ the ground-truth part label set\\
        \ \ \ \ \ \ $\mathcal{L} = \{l_1,\cdots,l_N\}$, where $N=H \cdot W$;\\
        \ the adversarial part label set $\mathcal{L'} = \{l_1',...,l_N'\}$;\\
        \ $\mathbf{x}$'s target location set $\mathcal{T} = \{t_1,..,t_N\}$;\\
        \ the maximal iteration steps $M$;\\
        \ the attack intensity $\epsilon$;\\
        \ the perturbation step size $\alpha$. \\
     }
     \Output{
        adversarial example $\mathbf{x'}$. \\
     }
     $\mathbf{x_0} \leftarrow \mathbf{x}, \mathbf{r} \leftarrow 0, m \leftarrow 0$\;

    \While{$m < M$}{

    $\mathbf{r_m} \leftarrow \sum_{t_n \in \mathcal{T}}[\nabla_{\mathbf{x_m}}F_{l_n'}(\mathbf{x_m},t_n)-\nabla_{\mathbf{x_m}}F_{l_n}(\mathbf{x_m},t_n)]$\;
        $\mathbf{r_m} \leftarrow \alpha \cdot \mathrm{sgn}(\mathbf{r_m})$\;
        $\mathbf{r} \leftarrow \mathrm{clip}(\mathbf{x_m} + \mathbf{r_m}-\mathbf{x}, \mathrm{min}=-\epsilon, \mathrm{max}=\epsilon)$\;
        $\mathbf{x_{m+1}} \leftarrow \mathrm{clip}(\mathbf{x}+\mathbf{r},\mathrm{min}=0, \mathrm{max}=1)$\;
        $m \leftarrow m+1$\;
    }
    $\mathbf{x'} \leftarrow \mathbf{x_{m}}$.

    \Return{
    $\mathbf{x'}$ \\
    }
\end{algorithm}

Since the first stage of ROCK is segmentation, and the score is highly dependent on segmentation results, one intuitive idea is to attack the segmentation network and force it to give false segmentation results. So we adapted the DAG \cite{dag} method to attack our model. DAG is originally an iterative targeted attack that tries to make the segmentation outputs to the specified adversarial labels. We empirically found that better attack results could be achieved if we changed the original DAG, which only attacked the pixels with ground-truth segmentation results per step, to attack all pixels simultaneously. We revised the DAG method to make it have a stronger attack performance and adjusted it to the $l_\infty$ setting. The modified version is shown in \cref{algo:algo1}. Here $F(\mathbf{x},t) \in \mathbb{R}^{K+1}$ denotes the logits at location $t$ of the image $\mathbf{x}$, and $F_{\rm foot}(\cdot,\cdot) \in \mathbb{R}$ denotes the value of part channel ${\rm foot}$, where $K$ denotes the number of parts (channels). $\mathcal{L}$ denotes the ground-truth part label set and $\mathcal{L'}$  denotes the adversarial part label set. The attack method's performance relied on the choice of $\mathcal{L'}$. We first designed four different variants of adaptive attack by setting different adversarial part labels $\mathcal{L'}$: (\romannumeral1) We removed adversarial part label set and changed DAG to be an untargeted attack. (\romannumeral2) We set $\mathcal{L'}$ to be ground-truth part labels $\mathcal{L}$ of another image, e.g., we tried to change the segmentation result of a \emph{bird} to that of a \emph{cat}. (\romannumeral3) We set $\mathcal{L'}$ as the background (zero) so that none of the parts could be segmented. (\romannumeral4) We randomly set $\mathcal{L'}$ for each pixel so that segmentation results were broken and none of the MCCs could be detected. We name these four variants as \emph{untargeted}, \emph{targeted}, \emph{background}, and \emph{random}, respectively. The idea of the fifth attack method is inspired by a previous work \cite{adaptive20}: we directly ignored the non-differentiable knowledge matching operation and only attacked the remaining differentiable computation of the final score. This attack followed the computation in \cref{eq:s3} so it took into account the importance of different linkage rules. We name this attack method as \emph{importance}.

%

All of these adaptive attacks were performed iteratively with PGD-40 with the perturbation step size $\epsilon/8$. \cref{tab:AttackROCK} shows that ROCK could defend against all five adaptive attacks relatively well. We report the \textbf{lowest} recognition accuracy corresponding to the strongest adaptive attack under the five adaptive attacks for each experimental setting in \cref{tab:facebase}. The accuracies of ROW models under white-box attacks using PGD-40 are listed alongside. The results show that the lowest accuracy of ROCK was higher than the accuracies of ROW models in each setting, indicating higher robustness of ROCK than that of ROW models under white-box attack settings.

\begin{figure}[!t]
    \centering
    \includegraphics[width=0.95\linewidth]{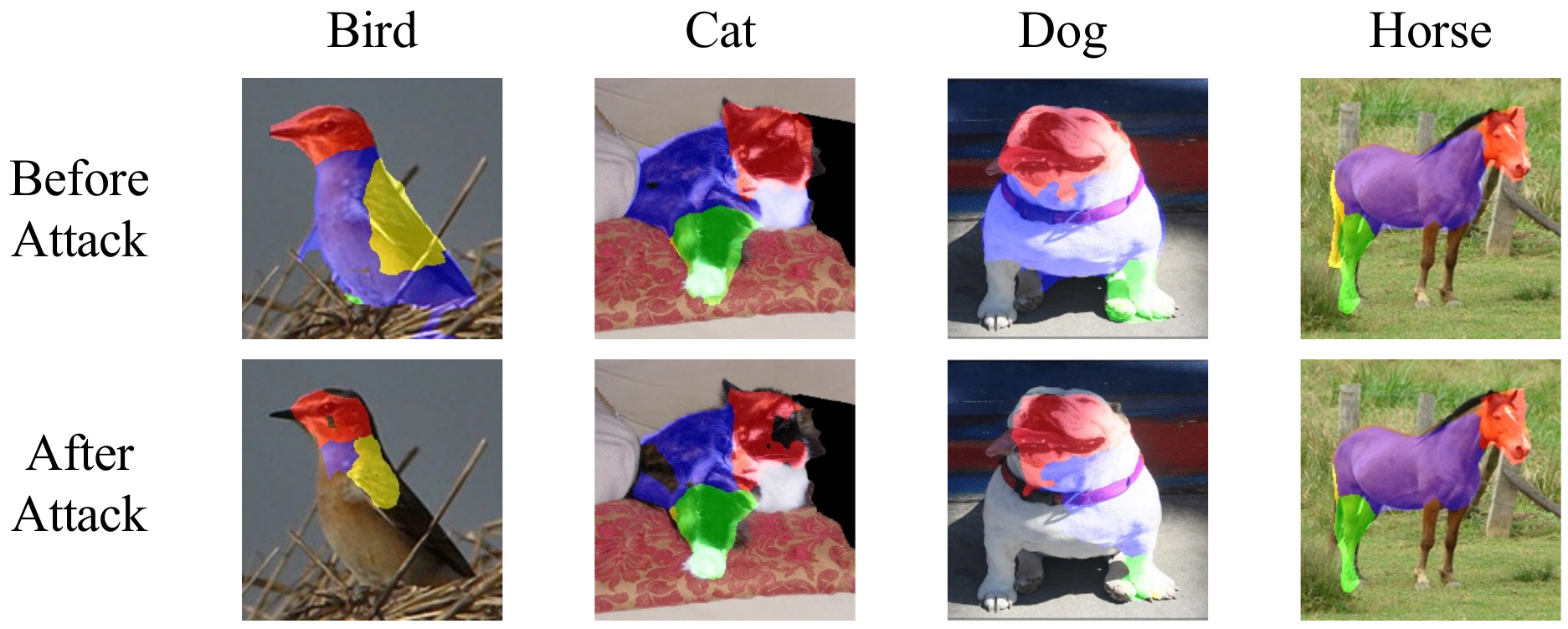}
    \caption{\xiaoli{ROCK's part segmentation results on examples which ROCK can still successfully identify after being attacked. The examples are chosen from Pascal-Part-C and the attack method was PGD-40 ($\epsilon = 1$).}}
    \label{fig:rebuttal_example}
\end{figure}

We then evaluated the robustness of ROCK and ROW models under AutoAttack \cite{AA}, a strong attack approach widely used in the community. It is a combination of variants of PGD \cite{AA} and \xiaoli{Square attack \cite{square}}, a query-based attack method. Since the three variants of PGD depend on the gradient of the target model whereas ROCK is not differentiable, we let them use the gradient used in the strongest adaptive attack. \cref{tab:facebase} shows that ROCK and ROW models had similar accuracies on benign images. However, when $\epsilon$ increased, the accuracies of ROW models dropped quickly, while ROCK maintained relatively high accuracy under both adaptive attacks and AutoAttack. \xiaoli{\cref{fig:rebuttal_example} shows ROCK's part segmentation results on some examples which ROCK can still successfully identify after being attacked. Although the segmentation results were significantly disturbed, some linkage relationships remained.} 

Since Face-Body is a toy dataset with only two categories, further experiments were only performed on Pascal-Part-C and PartImageNet-C.

\subsubsection{Black-Box Query-Based Attacks}
We then performed three query-based attacks, SPSA \cite{SPSA}, NES \cite{NES}, \xiaoli{and RayS \cite{RayS}}, to avoid the potential sub-optimal adaptive strategies. SPSA and NES can bypass the non-differentiable part of ROCK and obtain an approximate gradient of the final score by drawing random samples. SPSA samples direction vectors from the Rademacher distribution while NES samples from the Gaussian distribution. \xiaoli{RayS is a search-based method designed for hard-label outputs.} \cref{tab:query} shows that ROCK was consistently more robust than ROW models under various query-based attack methods and intensities. When the attack intensity $\epsilon$ was set to 8, ROCK was significantly more robust than ROW models. For instance, ROCK achieved  52.3\% accuracy while ROW$\rm{_X}$-101 achieved only 9.8\% accuracy on Pascale-Part-C under SPSA.




\subsection{Combining with Adversarial Training}
\xiaoli{ROCK can be combined with AT, which is generally recognized as an effective way to defend against adversarial attacks. We used different popular AT methods including PGD-AT \cite{pgd}, TRADES \cite{trades}, and TRADES-AWP \cite{AWP}. PGD-AT is directly trained on examples generated by PGD while TRADES tries to minimize the classification loss on benign images and the KL-divergence between outputs of benign and adversarial images. TRADES-AWP combines TRADES and AWP \cite{AWP} to further improve adversarial robustness. We notice that model weight averaging is helpful to improve robustness \cite{AWP_EMA}, so TRADES-AWP combined with exponential moving average, denoted as TRADES-AWP-EMA, was also evaluated to investigate the gains obtained by ROCK with strong AT methods. In what follows, We call TRADES, TRADES-AWP, and TRADES-AWP-EMA as TRADES-like methods.}

\begin{table*}[!t]
  \centering
  \caption{\xiaoli{Classification accuracies (\%) of ROW models and ROCK combined with four AT methods. The lowest accuracies of ROCK under five adaptive attacks are reported here.}}
  \small
  \setlength{\tabcolsep}{2pt}
  {
    \begin{tabular}{c|c||c|cccc||c|cccc}
    \toprule
    \multirow{3}*{\textbf{AT Method}}& \multirow{3}*{\textbf{Target}} & \multicolumn{5}{c||}{\textbf{Pascal-Part-C}} & \multicolumn{5}{c}{\textbf{PartImageNet-C}} \\
    \cmidrule{3-12}
                                    &                                & \textbf{Benign} & \multicolumn{2}{c}{\textbf{PGD-40}} & \multicolumn{2}{c||}{\textbf{AutoAttack}} & \textbf{Benign} & \multicolumn{2}{c}{\textbf{PGD-40}} & \multicolumn{2}{c}{\textbf{AutoAttack}} \\
                                    &                                & \textbf{Acc} & \textbf{$\bm{\epsilon =4}$} & \textbf{$\bm{\epsilon =8}$} & \textbf{$\bm{\epsilon =4}$} & \textbf{$\bm{\epsilon =8}$} & \textbf{Acc} & \textbf{$\bm{\epsilon =4}$} & \textbf{$\bm{\epsilon =8}$} & \textbf{$\bm{\epsilon =4}$} & \textbf{$\bm{\epsilon =8}$}\\
    \midrule
    \multirow{4}*{PGD-AT} & ROW-50 & $\bm{72.5}$ & $\bm{50.6}$ & $27.5$ & $\bm{49.5}$ & $25.4$ & $49.6$ & $26.6$ & $12.1$ & $25.5$ & $10.5$\\
                            & ROW$\rm{_M}$-50 & $70.4$ & $49.2$ & $26.1$ & $48.6$ & $25.5$ & $50.9$ & $29.5$ & $12.2$ & $27.0$ & $10.4$\\
                            & ROW$\rm{_X}$-101 & $71.9$ & $48.9$ & $26.6$ & $47.9$ & $25.3$ & $\bm{53.2}$ & $26.6$ & $11.2$ & $25.1$ & $10.5$ \\
     \cmidrule{2-12}
                            & ROCK & $71.5$ & $50.0$ & $\bm{29.0}$ & $49.0$ & $\bm{26.8}$ & $51.6$ & $\bm{31.1}$ & $\bm{14.6}$ & $\bm{28.1}$ & $\bm{11.8}$\\

    \midrule
    \multirow{4}*{TRADES} & ROW-50 & $70.3$ & $46.7$ & $27.9$ & $45.6$ & $25.8$ & $50.1$ & $27.2$ & $12.4$ & $23.8$ & $10.4$\\
                           & ROW$\rm{_M}$-50 & $73.7$ & $49.3$ & $27.1$ & $46.9$ & $25.2$ & $49.3$ & $26.3$ & $10.8$ & $22.8$ & $8.1$ \\
                            & ROW$\rm{_X}$-101 & $\bm{74.4}$ & $50.6$ & $27.9$ & $46.4$ & $25.9$ & $51.6$ & $30.5$ & $14.9$ & $28.1$ & $12.9$\\
     \cmidrule{2-12}
                           & ROCK & $70.4$ & $\bm{50.9}$ & $\bm{30.8}$ & $\bm{47.1}$ & $\bm{27.6}$ & $\bm{52.8}$ & $\bm{34.1}$ & $\bm{19.3}$ & $\bm{30.0}$ & $\bm{15.5}$\\
    \midrule
    \multirow{4}*{\xiaoli{TRADES-AWP}} & ROW-50 & $70.6$ & $47.2$ & $27.8$ & $45.9$  & $24.7$ & $52.3$ & $30.8$ & $14.3$ & $26.9$ & $11.1$ \\
                           & ROW$\rm{_M}$-50 & $70.2$ & $48.4$ & $28.9$ & $46.9$ & $25.6$ & $54.2$ & $31.9$ & $14.1$ & $27.0$ & $11.1$ \\
                            & ROW$\rm{_X}$-101 & $\bm{73.3}$ & $52.0$ & $30.3$ &$50.0$& $29.1$ & $54.6$ & $33.7$ & $16.1$ & $29.8$ & $13.6$ \\
     \cmidrule{2-12}
                           & ROCK & $73.0$ & $\bm{53.8}$ & $\bm{32.6}$ & $\bm{52.4}$ & $\bm{31.6}$ & $\bm{54.9}$ & $\bm{35.7}$ & $\bm{18.6}$ & $\bm{34.0}$ & $\bm{17.6}$ \\
    \midrule
    \multirow{5}*{\xiaoli{TRADES-AWP-EMA}} & ROW-50 & $70.9$ & $48.7$ & $28.8$ & $47.7$ & $26.3$ & $52.6$ & $31.3$ & $14.1$ & $26.6$  & $11.5$ \\
                           & ROW$\rm{_M}$-50 & $72.4$ & $50.6$ & $26.6$ & $49.8$ & $25.1$ & $53.5$ & $30.3$ & $13.4$ & $24.6$ & $9.7$ \\
                            & ROW$\rm{_X}$-101 & $75.1$ & $51.2$ & $30.3$ & $50.9$ & $28.8$ & $55.0$ & $33.4$ & $16.4$ & $29.3$ & $14.2$ \\
     \cmidrule{2-12}
                           & ROCK &$73.4$ & $53.4$ & $32.6$ & $52.6$ & $31.7$ & $54.5$ & $35.6$ & $18.5$ & $34.2$ & $17.3$ \\
                           & ROCK$\rm{_X}$-101 & $\bm{75.4}$ & $\bm{56.4}$ & $\bm{34.5}$ & $\bm{56.0}$ & $\bm{33.1}$ & $\bm{55.6}$ & $\bm{38.2}$ & $\bm{22.2}$ & $\bm{36.5}$ & $\bm{20.4}$ \\
    \bottomrule
    \end{tabular}
     }
  \label{tab:vocadv}
\end{table*}

\begin{table*}[!t]
  \centering
  \caption{\xiaoli{Classification accuracies (\%) on 15 types of image corruptions. ``mCA'' denotes the mean accuracy averaged over different types of corruptions. 
  The first block (row 2-5) lists the results of models with standard training, and the second block (row 6-10) lists the results of models with TRADES-AWP-EMA training.}}
  \small
  \resizebox{\linewidth}{!}{
  \setlength{\tabcolsep}{2pt}
  {
    \begin{tabular}{c|c|ccc|cccc|cccc|cccc}
    \toprule
    \multirow{2}*{\textbf{Method}}  & \multirow{2}*{\textbf{mCA}} & \multicolumn{3}{c|}{\textbf{Noise}}   & \multicolumn{4}{c|}{\textbf{Blur}} & \multicolumn{4}{c|}{\textbf{Weather}}  & \multicolumn{4}{c}{\textbf{Digital}}\\
                                     &                            & \textbf{Gaussian}  & \textbf{Shot} & \multicolumn{1}{c|}{\textbf{Impulse}}  & \textbf{Defocus}  & \textbf{Glass}  & \textbf{Motion} & \multicolumn{1}{c|}{\textbf{Zoom}}  & \textbf{Snow} & \textbf{Frost} & \textbf{Fog} & \multicolumn{1}{c|}{\textbf{Bright}}  & \textbf{Contra}  & \textbf{Elastic}  & \textbf{Pixel}  & \textbf{JPEG}\\
    \midrule
    \midrule
    ROW-50 &$35.8$&$28.0$&$28.4$&$25.1$&$36.9$&$42.0$&$42.3$&$31.1$&$21.2$&$24.2$&$24.8$&$45.2$&$16.7$&$57.7$&$58.0$&$55.8$ \\
    ROW$\rm{_M}$-50 &$38.2$&$30.7$&$29.7$&$27.4$&$39.1$&$43.8$&$42.2$&$\bm{31.7}$&$22.5$&$29.3$&$28.2$&$48.2$&$18.9$&$60.7$&$61.2$&$59.1$ \\
    ROW$\rm{_X}$-101 &$38.8$&$30.1$&$29.8$&$27.7$&$38.7$&$45.6$&$\bm{44.4}$&$30.9$&$25.9$&$29.4$&$28.5$&$50.5$&$17.4$&$61.5$&$61.7$&$59.5$ \\
    ROCK &$\bm{40.4}$&$\bm{31.3}$&$\bm{31.0}$&$\bm{29.3}$&$\bm{40.0}$&$\bm{48.6}$&$40.9$&$28.2$&$\bm{27.2}$&$\bm{32.6}$&$\bm{32.7}$&$\bm{55.1}$&$\bm{22.6}$&$\bm{63.1}$&$\bm{63.0}$&$\bm{61.0}$ \\
    \midrule
    \midrule
    ROW-50 & 29.1 & 25.1 & 28.3 & 21.8 & 29.7 & 37.1 & 36.0 & 32.8 & 18.1 & 15.1 & 3.0 & 34.5 & 4.6 & 50.3 & 48.8 & 51.0\\
    ROW$\rm{_M}$-50 & 29.1 & 29.7 & 27.2 & 20.5 & 28.0 & 36.2 & 35.0 & 30.5 & 18.4 & 16.5 & 3.1 & 34.3 & 4.4 & 50.7 & 49.8 & 51.7\\
    ROW$\rm{_X}$-101 & 30.3 & 32.1 & 29.9 & 21.7 & 28.9 & 37.0 & 36.3 & 32.2 & 19.6 & 17.3 & 2.7 & 36.6 & 4.6 & 52.2 & 51.1 & 53.0\\
    ROCK & $30.6$ & $34.8$ & $33.0$ & $\bm{25.8}$ & 27.2 & 34.5 & 33.6 & 30.1 & 23.3 & 17.6 & $\bm{3.1}$ & 36.6 & $\bm{4.7}$ & 51.9 & 49.9 & 53.5\\
    ROCK$\rm{_X}$-101 & $\bm{32.1}$ & $\bm{35.2}$ & $\bm{33.6}$ & 25.3 & $\bm{30.3}$ & $\bm{37.9}$ & $\bm{36.8}$ & $\bm{33.6}$ & $\bm{24.3}$ & $\bm{19.5}$ & 2.9 & $\bm{39.5}$ & 3.7 & $\bm{52.8}$ & $\bm{51.8}$ & $\bm{54.5}$ \\
    \bottomrule
    \end{tabular}
     }
     }
  \label{tab:corruption}%
\end{table*}%

We performed AT on ROW models by using adversarial examples generated by PGD and evaluated these models under PGD attack and AutoAttack. We performed AT on ROCK by using adversarial examples generated by the \emph{random} attack as described before, and evaluated ROCK under the five adaptive attacks and AutoAttack. The accuracy of ROCK could increase further if the same attack method is used in training and evaluation. 

For each AT method, we set the maximal perturbation $\epsilon$ to be 8 and extended the training to 300 epochs. On Pascal-Part-C, we set the step size of PGD iterations to be 1 and the number of iterative steps to be 10. On PartImageNet-C, we set the step size to be 2 and the number of iterative steps to be 5 to save the computational cost. In addition, we set the regularization parameter $1/\lambda$ to be 6 \xiaoli{for all  TRADES-like methods and $\gamma = 5 \times 10^{-3}$ for all AWP methods} on both datasets. When combining with TRADES-like methods, we found that ROCK had a bias to segment pixels to be \emph{background} with the increasing number of categories, which was harmful to the recognition accuracy. Therefore, we divided $\mathbf{B}(i,j)$ in \cref{eq:fb} by a scale factor $\sqrt{C}$ when combining ROCK with TRADES-like methods in inference, where $C$ is the number of object categories.

\xiaoli{Experimental results in \cref{tab:vocadv} and \cref{tab:AttackROCKAT} show that in general ROCK with AT was substantially more robust than ROW models with AT in different AT settings. For example, even with the strong AT method TRADES-AWP-EMA, ROCK could obtain 17.3\% accuracy under AutoAttack ($\epsilon = 8$) on PartImageNet-C, which was 5.8\% higher than ROW-50. The largest model ROW$\rm{_X}$-101 obtained the best robustness in general among different ROW models. We were interested in whether ROCK could be further boosted by using such a large backbone with strong AT method. So we replaced ROCK's backbone with a slightly modified ResNeXt-101 (it used two dilated convolutions like ROW$\rm{_M}$-50), denoted as ROCK$\rm{_X}$-101. When combined with TRADES-AWP-EMA, ROCK$\rm{_X}$-101 could further improve the robustness on both datasets (see the last row in \cref{tab:vocadv}). These results indicate that the combination of ROCK and AT can lead to better adversarial robustness.}

\subsection{Against Common Corruptions}
\xiaoli{We also investigated the robustness of ROCK under common image corruptions. Following the method proposed in \cite{imagenetc}, we generated different types of corrupted images on PartImageNet-C. The results of different models on these corrupted images are shown in \cref{tab:corruption}. Here the classification accuracies were tested under each corruption averaged on five levels of severity. We adopt accuracy as the metric to be consistent with other results, while this metric can be easily converted into the corruption error metric \cite{athe}. We report the results with both standard training setting and AT setting in \cref{tab:corruption}. In general ROCK had better robustness in most types of image corruptions in various training settings.}

\begin{table}[!t]
  \centering
  \caption{Classification accuracies (\%) of ROCK and its incomplete counterparts under PGD-40 attack. ``P'', ``K'', and ``W'' denote ``Part'', ``Knowledge'', and ``Weight'', respectively.}
  \small
  \setlength{\tabcolsep}{1.5pt}{
  \begin{tabular}{c||c|cc||c|cc}
    \toprule
    \multirow{2}*{\textbf{Target}} & \multicolumn{3}{c||}{\textbf{Pascal-Part-C}} & \multicolumn{3}{c}{\textbf{PartImageNet-C}}  \\
    \cmidrule{2-7}
                                 & \textbf{Benign} & \textbf{$\bm{\epsilon =1}$} & \textbf{$\bm{\epsilon =2}$}  & \textbf{Benign} & \textbf{$\bm{\epsilon =0.5}$} & \textbf{$\bm{\epsilon =1}$} \\
    \midrule
     w/o P & $82.2$ & $14.4$ & $2.8$ & $66.0$ & $24.9$ & $6.0$ \\
     w/ P, w/o K  & $\bm{84.8}$ & $19.4$ & $1.7$ & $\bm{69.8}$ & $27.9$ & $8.8$ \\
     w/ K, w/o W & $81.1$ & $27.5$ & $8.1$ & $66.5$ & $30.9$ & $10.7$ \\
     ROCK & $80.7$ & $\bm{28.2}$ & $\bm{8.6}$ & $66.7$ & $\bm{31.3}$ & $10.7$\\

    \bottomrule
    \end{tabular}
   }
  \label{tab:ab_abab}
\end{table}

\begin{table}[!t]
  \centering
  \caption{Classification accuracies (\%) of DeiT-Ti and ROCK with DeiT-Ti as backbone (ROCK$_{\rm{T}}$) under PGD-40 attack. DeiT-Ti was pretrained on ImageNet-1K and then fine-tuned on the two datasets for 20 epochs.}
  \small
  \setlength{\tabcolsep}{1.5pt}{
  \begin{tabular}{c||c|cc||c|cc}
    \toprule
    \multirow{2}*{\textbf{Target}} & \multicolumn{3}{c||}{\textbf{Pascal-Part-C}} & \multicolumn{3}{c}{\textbf{PartImageNet-C}}  \\
    \cmidrule{2-7}
                                 & \textbf{Benign} & \textbf{$\bm{\epsilon =1}$} & \textbf{$\bm{\epsilon =2}$}  & \textbf{Benign} & \textbf{$\bm{\epsilon =0.5}$} & \textbf{$\bm{\epsilon =1}$} \\
    \midrule
      DeiT-Ti & $91.9$ & $15.8$ & $0.1$ & $66.2$ & $24.1$ & $4.6$\\
	  ROCK$_{\rm{T}}$ & $89.4$ & $\bm{27.0}$ & $\bm{6.6}$ & $66.8$ & $\bm{30.2}$ & $\bm{8.1}$ \\
    \bottomrule
    \end{tabular}
   }
  \label{tab:ab_transformer}
\end{table}

\subsection{Ablation Study}
\label{sec:ablation}

\subsubsection{Effectiveness of Each Module}
We conducted experiments to investigate the usefulness of each module in ROCK. ROCK has two stages: part segmentation and knowledge matching. To show the effectiveness of part segmentation, we first evaluated a pure segmentation model, that only segmented out object label per pixel instead of part label (denoted by w/o P). The knowledge matching module is based on part segmentation results. To show the effectiveness of knowledge, we also evaluated part segmentation model that only uses the sum of confidence (output from softmax function) of all pixels (w/ P, w/o K) as the final score, without considering human prior knowledge. Finally, we evaluated the model that uses human prior knowledge but does not consider the importance of different knowledge (w/ K, w/o W). Experimental results in \cref{tab:ab_abab} showed that part labels themselves were effective in enhancing robustness, that the human prior knowledge based on part segmentation further enhanced robustness, and that taking the importance of different knowledge into account was slightly useful. We observed that human prior knowledge sightly decreased accuracies on benign images. The reason might be that human prior knowledge was not considered in the training stage of the part segmentation network.

\subsubsection{Transformer Backbone}
Recently Vision Transformers \cite{transformer, deit} have achieved impressive accuracy on object recognition tasks as well as several down-stream tasks. Transformers can also be used as the backbone of ROCK. We designed such a method in which coarse segmentation results were obtained naturally by merging all tokens of the transformer output. One 3 $\times$ 3 convolutional layer was added to adjust the segmentation channels. \cref{tab:ab_transformer} shows that ROCK using a Transformer DeiT-Ti \cite{deit} as the backbone was also more robust than the original DeiT-Ti.

\begin{table}[!t]
  \centering

  \caption{\xiaoli{Classification accuracies (\%) of ROCK$\rm{_{Few}}$ and comparison models under PGD-40 attack on the randomly selected 50 categories on PartImageNet-C. ROCK$\rm{_{Full}}$ denotes the model trained by all part labels.}}
  \small
  \setlength{\tabcolsep}{3.5pt}{
  \begin{tabular}{c||c|ccc}
    \toprule
    \multirow{1}*{\textbf{Target}} & \textbf{Benign}  & \textbf{$\bm{\epsilon =0.5}$} & \textbf{$\bm{\epsilon =1}$} & \textbf{$\bm{\epsilon =2}$} \\
    \midrule
     ROW-50 &  $72.8$ & $25.1$ & $5.4$ & $0.6$\\
     ROW$\rm{_M}$-50 & $71.8$ & $25.8$ & $6.3$ & $0.6$ \\
     ROW$\rm{_X}$-101 & $72.1$ & $22.8$ & $5.2$ & $0.0$\\
     \cmidrule{1-5}
     ROCK$\rm{_{Few}}$ & $71.1$ & $36.2$ & $14.5$ & $2.1$ \\
     ROCK$\rm{_{Full}}$ & $74.5$ & $43.2$ & $19.4$ & $3.9$ \\
    \bottomrule
    \end{tabular}
   }
  \label{tab:fewshot}
\end{table}

\subsection{Reducing the Dependence on Part Labels}
\label{sec:reduce}

\xiaoli{ROCK is fully-supervised so part labels are needed, which potentially influences its practical usefulness as 
getting part labels for an object is more expensive than getting a category label. We tried to alleviate this problem in two ways. }

\subsubsection{A Few-Shot Learning Setting}
\xiaoli{We first investigated the few-shot part label setting. We assume that only a few images of some categories have part labels, while most images of these categories do not have part labels and only have  category label. We randomly selected 50 categories on PartImageNet-C (125 categories in all) and randomly selected 10 images per category, then on these categories, we used the part labels of the selected 10 images only and set the part labels of other images invisible. We used all part labels of the remaining 75 categories to ensure that the part segmentation network have enough part label supervision. This condition could be relaxed if a large dataset with part labels is available for pre-training ROCK, but investigating that will deviate from the main concern of the paper, and we leave it as future work. We trained a part segmentation network on all 125 categories with these part labels only. We denote this model as ROCK$\rm{_{Few}}$, indicating that it used only 10 part labels per category on the 50 categories. 

To make use of the category labels of those images that lacked part labels, ROCK$\rm{_{Few}}$ used a simple training strategy to align the supervision of part labels and category labels. The alignment is based on such knowledge: for an image of a bird, although lacking part labels, we could be sure that its part labels should be composed of \textit{bird torso}, \textit{bird head}, $\cdots$, and \textit{background}, instead of part labels of other categories, e.g., \textit{dog torso}, \textit{dog head}, and so on. So on those images without part labels, we supervised ROCK$\rm{_{Few}}$  by random part labels of corresponding categories during the initial 100 training epochs, and by the part labels of corresponding categories predicted by ROCK$\rm{_{Few}}$ itself online during the last 100 training epochs. 

In \cref{tab:fewshot}, we report the results when only considering the classification of the selected 50 categories (the models were trained on 125 categories, but we masked out the remaining 75 categories). When performing attack on all 125 categories, ROCK$\rm{_{Few}}$ obtained 64.1\% benign accuracy, and 29.6\%, 8.2\% accuracies under PGD-40 attack when $\epsilon = 0.5$ and $1$, respectively. These results are only slightly lower than those with all labels for training (the three numbers are 66.7\%, 31.3\%, and 10.7\% in that case; see \cref{tab:facebase}). Therefore, ROCK has the ability to adaptively use these few-shot part labels to improve robustness.}


\begin{figure}[!t]
    \centering
    \includegraphics[width=0.85\linewidth]{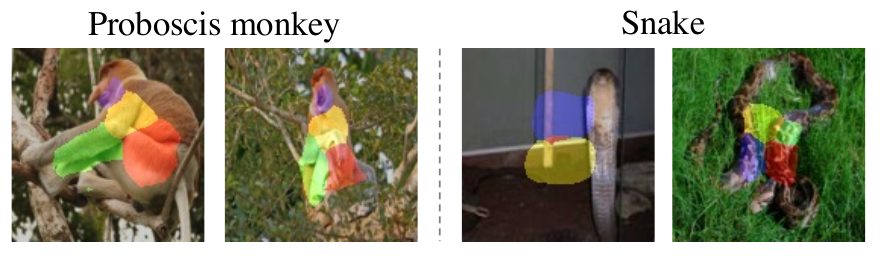}
    \vspace{-2mm}
    \caption{\xiaoli{Examples of pseudo part labels kept (left) and filtered out (right). }}
    \label{fig:pseudo}
\end{figure}

\begin{table}[!t]
  \centering

  \caption{\xiaoli{Classification accuracies (\%) of ROCK$\rm{_{Pseudo}}$ and some comparison models under PGD-40 attack when the models were trained on 45 selected categories on PartImageNet-C.}}
  \small
  \setlength{\tabcolsep}{3.5pt}{
  \begin{tabular}{c||c|ccc}
    \toprule
    \multirow{1}*{\textbf{Target}} & \textbf{Benign}  & \textbf{$\bm{\epsilon =0.5}$} & \textbf{$\bm{\epsilon =1}$} & \textbf{$\bm{\epsilon =2}$} \\
    \midrule
     ROW-50 &  $65.0$ & $27.3$ & $6.5$ & $0.1$ \\
     ROW$\rm{_M}$-50 & $67.3$ & $30.0$ & $7.3$ & $0.0$ \\
     ROW$\rm{_X}$-101 & $68.1$ & $29.6$ & $8.0$ & $0.0$\\
     \cmidrule{1-5}
     ROCK$\rm{_{Pseudo}}$ & $61.0$ & $35.0$ & $19.6$ & $4.3$ \\
     ROCK$\rm{_{Real}}$ & $69.1$ & $37.6$ & $22.6$ & $5.0$ \\
    \bottomrule
    \end{tabular}
   }
  \label{tab:pseudolabel}
\end{table}

\subsubsection{An Unsupervised Learning Setting}

\xiaoli{We also tried to use some unsupervised part segmentation methods to generate pseudo part labels, and then use these pseudo part labels to substitute real part labels for training ROCK. We investigated several unsupervised part segmentation techniques \cite{scops, aands, ganseg, groupvit}. Most methods need to train an independent model for the generation of pseudo part labels of a single category. We made a preliminary attempt on several categories on PartImageNet-C and chose one method \cite{aands} to conduct the complete experiments considering the balance between their training efficiency and result quality. This method generates part segmentation results by disentangling shape and appearance of an image in a self-supervised optimization way. Then the shape, represented by segmentation map, can be used as pseudo part labels.

With this method \cite{aands}, we trained 125 different unsupervised part segmentation models on the 125 categories on PartImageNet-C. However, the method failed to generate reasonable pseudo part labels on several categories on PartImageNet-C. For example, the models performed poorly on several categories of \textit{snakes} and the generated part labels were not even on the objects (\cref{fig:pseudo}). A potential reason is that PartImageNet-C is more difficult than the datasets used in the original paper \cite{aands}. We manually filtered out these categories and kept 45 categories with relatively reasonable pseudo part labels. \cref{fig:pseudo} shows some examples of pseudo part labels. As the pseudo part labels had no clear semantics compared with real part labels, such as \textit{head} and \textit{leg}, we cannot define linkage rules from human prior knowledge directly as illustrated in  \cref{fig:examplemerge}. Instead, we counted the linkage relationships between all pseudo part labels in the training dataset, and assumed that the linkage rule of a category should exist if 90\% images have such linkage relationships.

We trained a part segmentation model by the pseudo part labels on the 45 categories, denoted as ROCK$\rm{_{Pseudo}}$. For comparison, we also trained ROCK by the real part labels, denoted as ROCK$\rm{_{Real}}$, and ROW models on the 45 categories. The results are shown in \cref{tab:pseudolabel}. With the pseudo part labels, the robustness gains of ROCK$\rm{_{Pseudo}}$ are close to ROCK$\rm{_{Real}}$. ROCK$\rm{_{Pseudo}}$ caused a slight decrease in benign accuracy. This might be due to many misleading pseudo parts labels. 

Although the current unsupervised part segmentation methods cannot perform well on all categories, we believe that with the further advancement of unsupervised part segmentation, the gap between pseudo part labels and real part labels can be further reduced. Then the dependence of ROCK on real part labels can be largely alleviated.}

\begin{figure}[!t]
  \centering
  \includegraphics[width=0.89\linewidth]{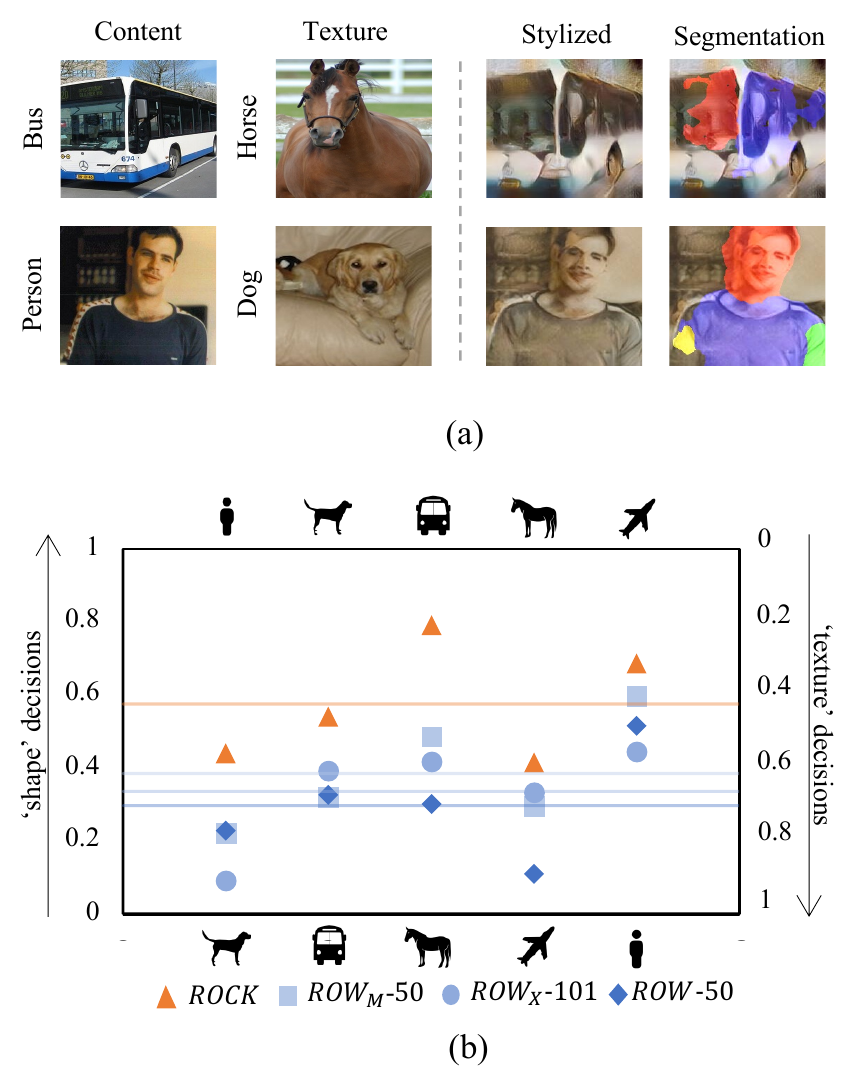}
  \vspace{-2mm}
  \caption{Texture and shape cue conflict experiment. (a) Two cue conflict images (the third column) and their segmentation results by ROCK (the fourth column). The first two columns show the images that were used to create the cue conflict images, e.g., we stylize \emph{bus} with \emph{horse}'s texture in the first row. (b) Classification results of the ROCK (orange) and ROW (blue) models on Pascal-Part-C. The top icons are the original categories (content images), while the bottom icons are the transferred categories (texture images). Compared with ROW models, ROCK biased more towards shape categories.}
  \label{fig:transfermerge}
\end{figure}

\subsection{Texture versus Shape Bias}
ROCK is inspired by the human recognition process. We were interested in whether its robustness aligns with human preference of shape bias over texture bias during object recognition \cite{shapetexture, shape}. To investigate whether ROCK depends more on shape-based features, following \cite{shapetexture}, we performed a cue conflict experiment that was based on images with contradicting texture and shape evidence. For example, we changed the texture of all \emph{person} images to the texture of \emph{dog} with the style transfer technique \cite{stransfer}, and then evaluated models on these images having shape of a \emph{person} and texture of a \emph{dog}. \cref{fig:transfermerge}(a) (row two, column three) shows one such example. Humans prefer to recognize it as a person \cite{shapetexture}. However, when we evaluated ROW models on such images with binary classification (only considering \emph{dog} and \emph{person}), 77.2\% images were recognized as \emph{dog} and only 22.8\% were recognized as \emph{person} by ROW-50. In contrast, ROCK recognized 54.1\% as \emph{dog} and 45.9\% as \emph{person} (column one in \cref{fig:transfermerge}(b)). We conducted five pairs of such experiments. \cref{fig:transfermerge}(b) shows that ROCK was much more biased towards the shape category than the three ROW models. The last column of \cref{fig:transfermerge}(a) shows some segmentation results using ROCK. More can be found in \cref{fig:appendix_style}. ROCK's segmentation can still output reasonable results on stylized images with significantly modified textures. The results demonstrate the robustness of ROCK with respect to texture change.

\section{Conclusion and Discussion}
\label{sec:diss}
In  this work, we propose ROCK, a novel attack-agnostic recognition model based on part segmentation, inspired by a well-known cognitive psychology theory recognition-by-components. By explicitly imitating human vision process, extensive experiments showed that ROCK is significantly more robust than conventional DNNs for object recognition across various attack settings. A cue conflict experiment demonstrated that ROCK’s decision-making process is more human-like. The success of ROCK highlights the great potential of part-based models and the advantages of incorporating human prior knowledge on robustness.

\subsection*{Limitations} 
\xiaoli{

First, ROCK requires part labels of objects, which are more expensive to obtain than the category labels. We consider the problem from the following aspects. (\romannumeral1) If the idea of ROCK based on recognition-by-components theory is proved to be much more robust than previous methods (which needs many researchers' effort), then it will urge the community to re-evaluate the cost for making ROCK and its extensions more practical, i.e., providing part labels of objects. This is possible especially in risk-averse applications such as autonomous driving because robustness is quite important in those applications. (\romannumeral2) Providing part labels is not that impractical. First, as we can see, many segmentation datasets are publicly available. Second, our experiments showed that for object recognition, part labels do not need to be fine-grained. $28 \times 28$ coarse part labels on images with the resolution of $224 \times 224$ were enough to achieve good results, making the annotation cost affordable. (\romannumeral3) As shown in \cref{sec:reduce}, we can figure out some methods to reduce the demand for part labels.


Second, currently ROCK is unable to handle objects without parts (e.g., a \emph{basketball}). Nevertheless, the notion of \textit{part} can be extended to meaningful patterns on objects (e.g., curves on a basketball). 

Finally, topological homogeneity used in this work is simple human prior knowledge about objects. More advanced knowledge, such as visual knowledge graph \cite{vkg} for each category, can be incorporated to improve the performance in future.
}

\ifCLASSOPTIONcompsoc
  \section*{Acknowledgments}
\else
  \section*{Acknowledgment}
\fi

This work was supported by the National Natural Science Foundation of China (Nos. U19B2034, 62061136001, 61836014).


\ifCLASSOPTIONcaptionsoff
  \newpage
\fi

\bibliographystyle{IEEEtran}
\bibliography{egbib.bib}

\begin{thebibliography}{10}
\providecommand{\url}[1]{#1}
\csname url@samestyle\endcsname
\providecommand{\newblock}{\relax}
\providecommand{\bibinfo}[2]{#2}
\providecommand{\BIBentrySTDinterwordspacing}{\spaceskip=0pt\relax}
\providecommand{\BIBentryALTinterwordstretchfactor}{4}
\providecommand{\BIBentryALTinterwordspacing}{\spaceskip=\fontdimen2\font plus
\BIBentryALTinterwordstretchfactor\fontdimen3\font minus
  \fontdimen4\font\relax}
\providecommand{\BIBforeignlanguage}[2]{{%
\expandafter\ifx\csname l@#1\endcsname\relax
\typeout{** WARNING: IEEEtran.bst: No hyphenation pattern has been}%
\typeout{** loaded for the language `#1'. Using the pattern for}%
\typeout{** the default language instead.}%
\else
\language=\csname l@#1\endcsname
\fi
#2}}
\providecommand{\BIBdecl}{\relax}
\BIBdecl

\bibitem{goodfellow14}
I.~J. Goodfellow, J.~Shlens, and C.~Szegedy, ``Explaining and harnessing
  adversarial examples,'' in \emph{ICLR}, 2015.

\bibitem{szegedy13}
C.~Szegedy, W.~Zaremba, I.~Sutskever, J.~Bruna, D.~Erhan, I.~Goodfellow, and
  R.~Fergus, ``Intriguing properties of neural networks,'' \emph{arXiv preprint
  arXiv:1312.6199}, 2013.

\bibitem{Eykholt18phy}
K.~Eykholt, I.~Evtimov, E.~Fernandes, B.~Li, A.~Rahmati, C.~Xiao, A.~Prakash,
  T.~Kohno, and D.~Song, ``Robust physical-world attacks on deep learning
  visual classification,'' in \emph{IEEE Conf. Comput. Vis. Pattern Recog.
  (CVPR)}, 2018, pp. 1625--1634.

\bibitem{zhu21}
X.~Zhu, X.~Li, J.~Li, Z.~Wang, and X.~Hu, ``Fooling thermal infrared pedestrian
  detectors in real world using small bulbs,'' in \emph{AAAI}, 2021, pp.
  3616--3624.

\bibitem{liao18}
F.~Liao, M.~Liang, Y.~Dong, T.~Pang, X.~Hu, and J.~Zhu, ``Defense against
  adversarial attacks using high-level representation guided denoiser,'' in
  \emph{IEEE Conf. Comput. Vis. Pattern Recog. (CVPR)}, 2018, pp. 1778--1787.

\bibitem{ZhouL021}
D.~Zhou, T.~Liu, B.~Han, N.~Wang, C.~Peng, and X.~Gao, ``Towards defending
  against adversarial examples via attack-invariant features,'' in \emph{Int.
  Conf. Mach. Learn. (ICML)}, vol. 139, 2021, pp. 12\,835--12\,845.

\bibitem{Pang18}
T.~Pang, C.~Du, and J.~Zhu, ``Max-mahalanobis linear discriminant analysis
  networks,'' in \emph{Int. Conf. Mach. Learn. (ICML)}, vol.~80, 2018, pp.
  4013--4022.

\bibitem{Pang20}
T.~Pang, K.~Xu, Y.~Dong, C.~Du, N.~Chen, and J.~Zhu, ``Rethinking softmax
  cross-entropy loss for adversarial robustness,'' in \emph{ICLR}, 2020.

\bibitem{RossD18}
A.~S. Ross and F.~Doshi{-}Velez, ``Improving the adversarial robustness and
  interpretability of deep neural networks by regularizing their input
  gradients,'' in \emph{AAAI}, 2018, pp. 1660--1669.

\bibitem{YeatsCL21}
E.~C. Yeats, Y.~Chen, and H.~Li, ``Improving gradient regularization using
  complex-valued neural networks,'' in \emph{Int. Conf. Mach. Learn. (ICML)},
  vol. 139, 2021, pp. 11\,953--11\,963.

\bibitem{AthalyeC018}
A.~Athalye, N.~Carlini, and D.~A. Wagner, ``Obfuscated gradients give a false
  sense of security: Circumventing defenses to adversarial examples,'' in
  \emph{Int. Conf. Mach. Learn. (ICML)}, vol.~80, 2018, pp. 274--283.

\bibitem{adaptive20}
F.~Tram{\`{e}}r, N.~Carlini, W.~Brendel, and A.~Madry, ``On adaptive attacks to
  adversarial example defenses,'' in \emph{Adv. Neural Inform. Process. Syst.
  (NeurIPS)}, 2020.

\bibitem{DongDP0020}
Y.~Dong, Z.~Deng, T.~Pang, J.~Zhu, and H.~Su, ``Adversarial distributional
  training for robust deep learning,'' in \emph{Adv. Neural Inform. Process.
  Syst. (NeurIPS)}, 2020.

\bibitem{gene}
S.~Gowal, S.-A. Rebuffi, O.~Wiles, F.~Stimberg, D.~A. Calian, and T.~A. Mann,
  ``Improving robustness using generated data,'' \emph{Adv. Neural Inform.
  Process. Syst. (NeurIPS)}, vol.~34, 2021.

\bibitem{athe}
T.~Pang, X.~Yang, Y.~Dong, T.~Xu, J.~Zhu, and H.~Su, ``Boosting adversarial
  training with hypersphere embedding,'' in \emph{Adv. Neural Inform. Process.
  Syst. (NeurIPS)}, 2020.

\bibitem{trades}
H.~Zhang, Y.~Yu, J.~Jiao, E.~P. Xing, L.~E. Ghaoui, and M.~I. Jordan,
  ``Theoretically principled trade-off between robustness and accuracy,'' in
  \emph{Int. Conf. Mach. Learn. (ICML)}, vol.~97, 2019, pp. 7472--7482.

\bibitem{AWP}
D.~Wu, S.~Xia, and Y.~Wang, ``Adversarial weight perturbation helps robust
  generalization,'' in \emph{Adv. Neural Inform. Process. Syst. (NeurIPS)},
  2020.

\bibitem{atg}
H.~Zhang, H.~Chen, Z.~Song, D.~S. Boning, I.~S. Dhillon, and C.~Hsieh, ``The
  limitations of adversarial training and the blind-spot attack,'' in
  \emph{ICLR}, 2019.

\bibitem{humanadv}
Z.~Zhou and C.~Firestone, ``Humans can decipher adversarial images,''
  \emph{Nature Communications}, vol.~10, no.~1, pp. 1--9, 2019.

\bibitem{imagenetp}
X.~Li, J.~Li, T.~Dai, J.~Shi, J.~Zhu, and X.~Hu, ``Rethinking natural
  adversarial examples for classification models,'' \emph{arXiv preprint
  arXiv:2102.11731}, 2021.

\bibitem{RBC}
I.~Biederman, ``Recognition-by-components: a theory of human image
  understanding.'' \emph{Psychological Review}, vol.~94, no.~2, p. 115, 1987.

\bibitem{attentioneye}
J.~E. Hoffman and B.~Subramaniam, ``The role of visual attention in saccadic
  eye movements,'' \emph{Perception \& Psychophysics}, vol.~57, no.~6, pp.
  787--795, 1995.

\bibitem{oldrbc}
E.~Rosch and C.~B. Mervis, ``Family resemblances: Studies in the internal
  structure of categories,'' \emph{Cognitive Psychology}, vol.~7, no.~4, pp.
  573--605, 1975.

\bibitem{opc}
B.~Tversky and K.~Hemenway, ``Objects, parts, and categories.'' \emph{Journal
  of Experimental Psychology: General}, vol. 113, no.~2, p. 169, 1984.

\bibitem{infants}
R.~A. Haaf, A.~L. Fulkerson, B.~J. Jablonski, J.~M. Hupp, S.~S. Shull, and
  L.~Pescara-Kovach, ``Object recognition and attention to object components by
  preschool children and 4-month-old infants,'' \emph{Journal of Experimental
  Child Psychology}, vol.~86, no.~2, pp. 108--123, 2003.

\bibitem{bugfeature}
A.~Ilyas, S.~Santurkar, D.~Tsipras, L.~Engstrom, B.~Tran, and A.~Madry,
  ``Adversarial examples are not bugs, they are features,'' in \emph{Adv.
  Neural Inform. Process. Syst. (NeurIPS)}, 2019, pp. 125--136.

\bibitem{shapetexture}
R.~Geirhos, P.~Rubisch, C.~Michaelis, M.~Bethge \emph{et~al.},
  ``Imagenet-trained cnns are biased towards texture; increasing shape bias
  improves accuracy and robustness,'' in \emph{ICLR}, 2019.

\bibitem{partpro}
M.~C. Burl, M.~Weber, and P.~Perona, ``A probabilistic approach to object
  recognition using local photometry and global geometry,'' in \emph{Eur. Conf.
  Comput. Vis. (ECCV)}, vol. 1407, 1998, pp. 628--641.

\bibitem{partsl}
D.~J. Crandall, P.~F. Felzenszwalb, and D.~P. Huttenlocher, ``Spatial priors
  for part-based recognition using statistical models,'' in \emph{IEEE Conf.
  Comput. Vis. Pattern Recog. (CVPR)}, 2005, pp. 10--17.

\bibitem{partdis}
P.~F. Felzenszwalb, R.~B. Girshick, D.~A. McAllester, and D.~Ramanan, ``Object
  detection with discriminatively trained part-based models,'' \emph{IEEE
  Trans. Pattern Anal. Mach. Intell. (TPAMI)}, vol.~32, no.~9, pp. 1627--1645,
  2010.

\bibitem{partlearning}
R.~Fergus, P.~Perona, and A.~Zisserman, ``Object class recognition by
  unsupervised scale-invariant learning,'' in \emph{IEEE Conf. Comput. Vis.
  Pattern Recog. (CVPR)}, 2003, pp. 264--271.

\bibitem{pgd}
A.~Madry, A.~Makelov, L.~Schmidt, D.~Tsipras, and A.~Vladu, ``Towards deep
  learning models resistant to adversarial attacks,'' in \emph{ICLR}, 2018.

\bibitem{concurrent}
C.~Sitawarin, K.~Pongmala, Y.~Chen, N.~Carlini, and D.~A. Wagner, ``Part-based
  models improve adversarial robustness,'' \emph{arXiv preprint
  arXiv:2209.09117}, vol. abs/2209.09117, 2022.

\bibitem{deepfool}
S.~Moosavi{-}Dezfooli, A.~Fawzi, and P.~Frossard, ``Deepfool: {A} simple and
  accurate method to fool deep neural networks,'' in \emph{IEEE Conf. Comput.
  Vis. Pattern Recog. (CVPR)}, Jun. 2016, pp. 2574--2582.

\bibitem{mim}
Y.~Dong, F.~Liao, T.~Pang, H.~Su, J.~Zhu, X.~Hu, and J.~Li, ``Boosting
  adversarial attacks with momentum,'' in \emph{IEEE Conf. Comput. Vis. Pattern
  Recog. (CVPR)}, 2018, pp. 9185--9193.

\bibitem{madry17cw}
A.~Madry, A.~Makelov, L.~Schmidt, D.~Tsipras, and A.~Vladu, ``Towards deep
  learning models resistant to adversarial attacks,'' \emph{arXiv preprint
  arXiv:1706.06083}, 2017.

\bibitem{benchmark}
Y.~Dong, Q.~Fu, X.~Yang, T.~Pang, H.~Su, Z.~Xiao, and J.~Zhu, ``Benchmarking
  adversarial robustness on image classification,'' in \emph{IEEE Conf. Comput.
  Vis. Pattern Recog. (CVPR)}, 2020, pp. 318--328.

\bibitem{ob1}
C.~Guo, M.~Rana, M.~Ciss{\'{e}}, and L.~van~der Maaten, ``Countering
  adversarial images using input transformations,'' in \emph{ICLR}, 2018.

\bibitem{ob2}
C.~Xie, J.~Wang, Z.~Zhang, Z.~Ren, and A.~L. Yuille, ``Mitigating adversarial
  effects through randomization,'' in \emph{ICLR}, 2018.

\bibitem{SPSA}
J.~Uesato, B.~O'Donoghue, P.~Kohli, and A.~van~den Oord, ``Adversarial risk and
  the dangers of evaluating against weak attacks,'' in \emph{Int. Conf. Mach.
  Learn. (ICML)}, vol.~80, 2018, pp. 5032--5041.

\bibitem{NES}
A.~Ilyas, L.~Engstrom, A.~Athalye, and J.~Lin, ``Black-box adversarial attacks
  with limited queries and information,'' in \emph{Int. Conf. Mach. Learn.
  (ICML)}, vol.~80, 2018, pp. 2142--2151.

\bibitem{dag}
C.~Xie, J.~Wang, Z.~Zhang, Y.~Zhou, L.~Xie, and A.~L. Yuille, ``Adversarial
  examples for semantic segmentation and object detection,'' in \emph{Int.
  Conf. Comput. Vis. (ICCV)}, 2017, pp. 1378--1387.

\bibitem{segadv}
V.~Fischer, M.~C. Kumar, J.~H. Metzen, and T.~Brox, ``Adversarial examples for
  semantic image segmentation,'' in \emph{{ICLR} (Workshop)}, 2017.

\bibitem{ArnabMT18}
A.~Arnab, O.~Miksik, and P.~H.~S. Torr, ``On the robustness of semantic
  segmentation models to adversarial attacks,'' in \emph{IEEE Conf. Comput.
  Vis. Pattern Recog. (CVPR)}, 2018, pp. 888--897.

\bibitem{character}
C.~Xiao, R.~Deng, B.~Li, F.~Yu, M.~Liu, and D.~Song, ``Characterizing
  adversarial examples based on spatial consistency information for semantic
  segmentation,'' in \emph{Eur. Conf. Comput. Vis. (ECCV)}, vol. 11214, 2018,
  pp. 220--237.

\bibitem{wavelet}
C.~Papageorgiou, M.~Oren, and T.~A. Poggio, ``A general framework for object
  detection,'' in \emph{Int. Conf. Comput. Vis. (ICCV)}, 1998, pp. 555--562.

\bibitem{sift}
D.~G. Lowe, ``Object recognition from local scale-invariant features,'' in
  \emph{Int. Conf. Comput. Vis. (ICCV)}, 1999, pp. 1150--1157.

\bibitem{compas}
J.~He, A.~Kortylewski, and A.~Yuille, ``Compas: Representation learning with
  compositional part sharing for few-shot classification,'' \emph{arXiv
  preprint arXiv:2101.11878}, 2021.

\bibitem{acn}
W.~Xu, H.~Wang, Z.~Tu \emph{et~al.}, ``Attentional constellation nets for
  few-shot learning,'' in \emph{ICLR}, 2020.

\bibitem{wsod}
D.~Zhang, W.~Zeng, J.~Yao, and J.~Han, ``Weakly supervised object detection
  using proposal- and semantic-level relationships,'' \emph{IEEE Trans. Pattern
  Anal. Mach. Intell. (TPAMI)}, vol.~44, no.~6, pp. 3349--3363, 2022.

\bibitem{lip}
K.~Gong, X.~Liang, D.~Zhang, X.~Shen, and L.~Lin, ``Look into person:
  Self-supervised structure-sensitive learning and a new benchmark for human
  parsing,'' \emph{IEEE Conf. Comput. Vis. Pattern Recog. (CVPR)}, pp.
  6757--6765, 2017.

\bibitem{Lee0W020}
C.~Lee, Z.~Liu, L.~Wu, and P.~Luo, ``Maskgan: Towards diverse and interactive
  facial image manipulation,'' in \emph{IEEE Conf. Comput. Vis. Pattern Recog.
  (CVPR)}, 2020, pp. 5548--5557.

\bibitem{GMNet}
U.~Michieli, E.~Borsato, L.~Rossi, and P.~Zanuttigh, ``Gmnet: Graph matching
  network for large scale part semantic segmentation in the wild,'' in
  \emph{Eur. Conf. Comput. Vis. (ECCV)}, vol. 12353, 2020, pp. 397--414.

\bibitem{bound}
Y.~Zhao, J.~Li, Y.~Zhang, and Y.~Tian, ``Multi-class part parsing with joint
  boundary-semantic awareness,'' in \emph{Int. Conf. Comput. Vis. (ICCV)},
  2019, pp. 9176--9185.

\bibitem{cgpart}
Q.~Liu, A.~Kortylewski, Z.~Zhang, Z.~Li, M.~Guo, Q.~Liu, X.~Yuan, J.~Mu,
  W.~Qiu, and A.~Yuille, ``Cgpart: A part segmentation dataset based on 3d
  computer graphics models,'' \emph{IEEE Conf. Comput. Vis. Pattern Recog.
  (CVPR)}, 2022.

\bibitem{partnet}
K.~Mo, S.~Zhu, A.~X. Chang, L.~Yi, S.~Tripathi, L.~J. Guibas, and H.~Su,
  ``Partnet: {A} large-scale benchmark for fine-grained and hierarchical
  part-level 3d object understanding,'' in \emph{IEEE Conf. Comput. Vis.
  Pattern Recog. (CVPR)}, 2019, pp. 909--918.

\bibitem{capsulesr}
P.~K. Mensah, A.~F. Adekoya, M.~A. Ayidzoe, and E.~Y. Baagyire, ``Capsule
  networks - {A} survey,'' \emph{J. King Saud Univ. Comput. Inf. Sci.},
  vol.~34, no.~1, pp. 1295--1310, 2022.

\bibitem{porvsr}
Y.~Liu, D.~Zhang, Q.~Zhang, and J.~Han, ``Part-object relational visual
  saliency,'' \emph{IEEE Trans. Pattern Anal. Mach. Intell. (TPAMI)}, vol.~44,
  no.~7, pp. 3688--3704, 2022.

\bibitem{onfocusr}
D.~Zhang, B.~Wang, G.~Wang, Q.~Zhang, J.~Zhang, J.~Han, and Z.~You, ``Onfocus
  detection: Identifying individual-camera eye contact from unconstrained
  images,'' \emph{Science China Information Sciences}, vol.~65, no.~6, pp.
  1--12, 2022.

\bibitem{imagenet}
O.~Russakovsky, J.~Deng, H.~Su, J.~Krause, S.~Satheesh, S.~Ma \emph{et~al.},
  ``Imagenet large scale visual recognition challenge,'' \emph{Int. J. Comput.
  Vis. (IJCV)}, vol. 115, no.~3, pp. 211--252, 2015.

\bibitem{capsulenor}
J.~Gu, V.~Tresp, and H.~Hu, ``Capsule network is not more robust than
  convolutional network,'' in \emph{IEEE Conf. Comput. Vis. Pattern Recog.
  (CVPR)}, 2021, pp. 14\,309--14\,317.

\bibitem{advcapsule}
F.~Michels, T.~Uelwer, E.~Upschulte, and S.~Harmeling, ``On the vulnerability
  of capsule networks to adversarial attacks,'' \emph{arXiv preprint
  arXiv:1906.03612}, 2019.

\bibitem{neuralsym}
K.~Yi, J.~Wu, C.~Gan, A.~Torralba, P.~Kohli, and J.~Tenenbaum,
  ``Neural-symbolic {VQA:} disentangling reasoning from vision and language
  understanding,'' in \emph{Adv. Neural Inform. Process. Syst. (NeurIPS)},
  2018, pp. 1039--1050.

\bibitem{informedml}
L.~von Rueden, S.~Mayer, K.~Beckh, B.~Georgiev \emph{et~al.}, ``Informed
  machine learning-a taxonomy and survey of integrating prior knowledge into
  learning systems,'' in \emph{IEEE Transactions on Knowledge and Data
  Engineering}, 2021.

\bibitem{KEMLP}
N.~M. G{\"{u}}rel, X.~Qi, L.~Rimanic, C.~Zhang, and B.~Li, ``Knowledge enhanced
  machine learning pipeline against diverse adversarial attacks,'' in
  \emph{Int. Conf. Mach. Learn. (ICML)}, vol. 139, 2021, pp. 3976--3987.

\bibitem{long2015fully}
J.~Long, E.~Shelhamer, and T.~Darrell, ``Fully convolutional networks for
  semantic segmentation,'' in \emph{IEEE Conf. Comput. Vis. Pattern Recog.
  (CVPR)}, 2015, pp. 3431--3440.

\bibitem{parttoc}
X.~Chen, R.~Mottaghi, X.~Liu, S.~Fidler, R.~Urtasun, and A.~L. Yuille, ``Detect
  what you can: Detecting and representing objects using holistic models and
  body parts,'' in \emph{IEEE Conf. Comput. Vis. Pattern Recog. (CVPR)}, 2014,
  pp. 1979--1986.

\bibitem{partimagenet}
J.~He, S.~Yang, S.~Yang, A.~Kortylewski, X.~Yuan, J.-N. Chen, S.~Liu, C.~Yang,
  and A.~Yuille, ``Partimagenet: A large, high-quality dataset of parts,''
  \emph{arXiv preprint arXiv:2112.00933}, 2021.

\bibitem{pascal2010}
M.~Everingham, L.~V. Gool, C.~K.~I. Williams, J.~M. Winn, and A.~Zisserman,
  ``The pascal visual object classes {(VOC)} challenge,'' \emph{Int. J. Comput.
  Vis. (IJCV)}, vol.~88, no.~2, pp. 303--338, 2010.

\bibitem{resnet}
K.~He, X.~Zhang, S.~Ren, and J.~Sun, ``Deep residual learning for image
  recognition,'' in \emph{IEEE Conf. Comput. Vis. Pattern Recog. (CVPR)}, 2016,
  pp. 770--778.

\bibitem{ResNeXt}
S.~Xie, R.~B. Girshick, P.~Doll{\'{a}}r, Z.~Tu, and K.~He, ``Aggregated
  residual transformations for deep neural networks,'' in \emph{IEEE Conf.
  Comput. Vis. Pattern Recog. (CVPR)}, 2017, pp. 5987--5995.

\bibitem{deepwise}
A.~G. Howard, M.~Zhu, B.~Chen, D.~Kalenichenko, W.~Wang, T.~Weyand,
  M.~Andreetto, and H.~Adam, ``Mobilenets: Efficient convolutional neural
  networks for mobile vision applications,'' \emph{arXiv preprint
  arXiv:1704.04861}, 2017.

\bibitem{deeplabv2}
L.~Chen, G.~Papandreou, I.~Kokkinos, K.~Murphy, and A.~L. Yuille, ``Deeplab:
  Semantic image segmentation with deep convolutional nets, atrous convolution,
  and fully connected crfs,'' \emph{IEEE Trans. Pattern Anal. Mach. Intell.
  (TPAMI)}, vol.~40, no.~4, pp. 834--848, 2018.

\bibitem{focal}
T.~Lin, P.~Goyal, R.~B. Girshick, K.~He, and P.~Doll{\'{a}}r, ``Focal loss for
  dense object detection,'' \emph{IEEE Trans. Pattern Anal. Mach. Intell.
  (TPAMI)}, vol.~42, no.~2, pp. 318--327, 2020.

\bibitem{fd}
C.~Xie, Y.~Wu, L.~van~der Maaten, A.~L. Yuille, and K.~He, ``Feature denoising
  for improving adversarial robustness,'' in \emph{IEEE Conf. Comput. Vis.
  Pattern Recog. (CVPR)}, 2019, pp. 501--509.

\bibitem{transfer}
N.~Papernot, P.~McDaniel, and I.~Goodfellow, ``Transferability in machine
  learning: from phenomena to black-box attacks using adversarial samples,''
  \emph{arXiv preprint arXiv:1605.07277}, 2016.

\bibitem{AA}
F.~Croce and M.~Hein, ``Reliable evaluation of adversarial robustness with an
  ensemble of diverse parameter-free attacks,'' in \emph{Int. Conf. Mach.
  Learn. (ICML)}, vol. 119, 2020, pp. 2206--2216.

\bibitem{square}
M.~Andriushchenko, F.~Croce, N.~Flammarion, and M.~Hein, ``Square attack: {A}
  query-efficient black-box adversarial attack via random search,'' in
  \emph{Eur. Conf. Comput. Vis. (ECCV)}, vol. 12368, 2020, pp. 484--501.

\bibitem{RayS}
J.~Chen and Q.~Gu, ``Rays: {A} ray searching method for hard-label adversarial
  attack.''\hskip 1em plus 0.5em minus 0.4em\relax {ACM}, 2020, pp. 1739--1747.

\bibitem{AWP_EMA}
S.~Rebuffi, S.~Gowal, D.~A. Calian, F.~Stimberg, O.~Wiles, and T.~A. Mann,
  ``Data augmentation can improve robustness,'' in \emph{Adv. Neural Inform.
  Process. Syst. (NeurIPS)}, 2021, pp. 29\,935--29\,948.

\bibitem{imagenetc}
D.~Hendrycks and T.~G. Dietterich, ``Benchmarking neural network robustness to
  common corruptions and perturbations,'' in \emph{ICLR}, 2019.

\bibitem{transformer}
A.~Dosovitskiy, L.~Beyer, A.~Kolesnikov, D.~Weissenborn \emph{et~al.}, ``An
  image is worth 16x16 words: Transformers for image recognition at scale,'' in
  \emph{ICLR}, 2021.

\bibitem{deit}
H.~Touvron, M.~Cord, M.~Douze \emph{et~al.}, ``Training data-efficient image
  transformers {\&} distillation through attention,'' in \emph{Int. Conf. Mach.
  Learn. (ICML)}, vol. 139, 2021, pp. 10\,347--10\,357.

\bibitem{scops}
W.~Hung, V.~Jampani, S.~Liu, P.~Molchanov, M.~Yang, and J.~Kautz, ``{SCOPS:}
  self-supervised co-part segmentation,'' in \emph{IEEE Conf. Comput. Vis.
  Pattern Recog. (CVPR)}, 2019, pp. 869--878.

\bibitem{aands}
S.~Liu, L.~Zhang, X.~Yang, H.~Su, and J.~Zhu, ``Unsupervised part segmentation
  through disentangling appearance and shape,'' in \emph{IEEE Conf. Comput.
  Vis. Pattern Recog. (CVPR)}, 2021, pp. 8355--8364.

\bibitem{ganseg}
X.~He, B.~Wandt, and H.~Rhodin, ``Ganseg: Learning to segment by unsupervised
  hierarchical image generation,'' in \emph{IEEE Conf. Comput. Vis. Pattern
  Recog. (CVPR)}, 2022, pp. 1215--1225.

\bibitem{groupvit}
J.~Xu, S.~D. Mello, S.~Liu, W.~Byeon, T.~M. Breuel, J.~Kautz, and X.~Wang,
  ``Groupvit: Semantic segmentation emerges from text supervision,'' in
  \emph{IEEE Conf. Comput. Vis. Pattern Recog. (CVPR)}, 2022, pp.
  18\,113--18\,123.

\bibitem{shape}
M.~Sun, Z.~Li, C.~Xiao, H.~Qiu, B.~Kailkhura, M.~Liu, and B.~Li, ``Can shape
  structure features improve model robustness under diverse adversarial
  settings?'' in \emph{Int. Conf. Comput. Vis. (ICCV)}, 2021, pp. 7526--7535.

\bibitem{stransfer}
X.~Huang and S.~J. Belongie, ``Arbitrary style transfer in real-time with
  adaptive instance normalization,'' in \emph{Int. Conf. Comput. Vis. (ICCV)},
  2017, pp. 1510--1519.

\bibitem{vkg}
S.~Monka, L.~Halilaj, S.~Schmid, and A.~Rettinger, ``Learning visual models
  using a knowledge graph as a trainer,'' in \emph{ISWC}, 2021, pp. 357--373.

\end{thebibliography}

\begin{IEEEbiography}[{\includegraphics[width=1in,height=1.25in,clip,keepaspectratio]{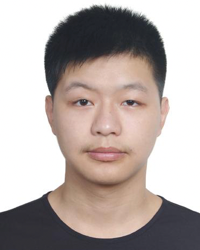}}]{Xiao Li} received the B.E. degree from Tsinghua University, China, in 2020, where he is currently pursuing the Ph.D. degree with the Department of Computer Science and Technology. His research interests include robustness and privacy of machine learning systems and object detection, especially making deep neural networks transparent and robust against adversarial attacks.
\end{IEEEbiography}

\begin{IEEEbiography}[{\includegraphics[width=1in,height=1.25in,clip,keepaspectratio]{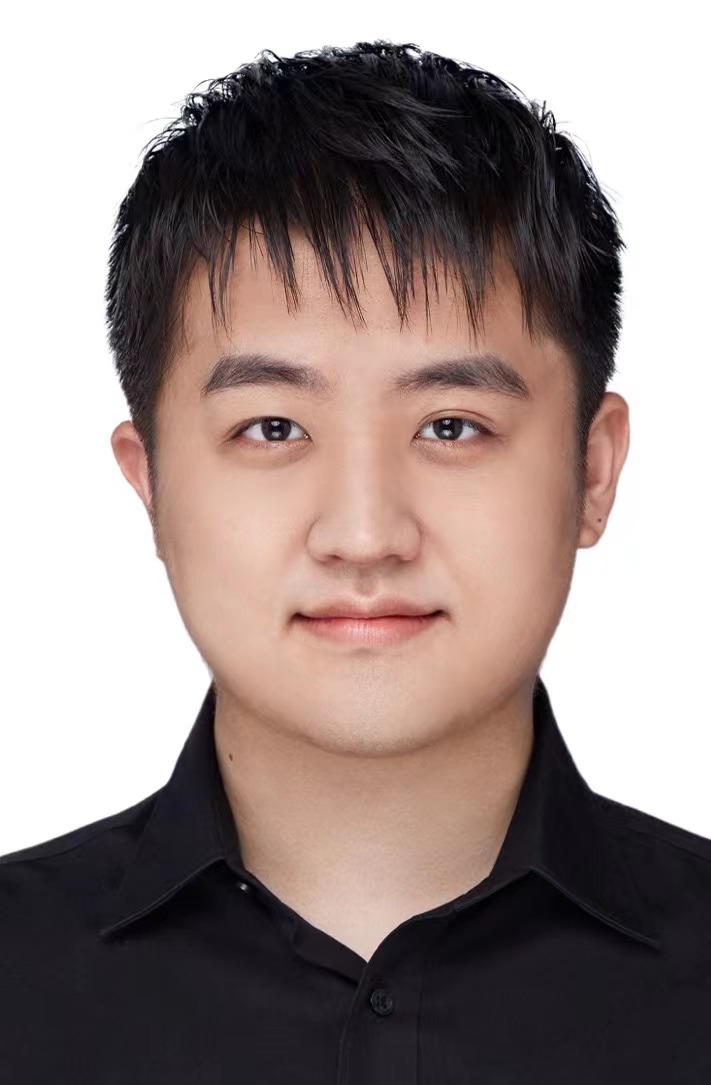}}]{Ziqi Wang} received the B.E. degree from the Department of Computer Science and Technology, Tsinghua University, in 2021. He is currently a Ph.D. student at the University of Illinois Urbana-Champaign, focusing on natural language processing and machine learning. His primary interest is information extraction and knowledge grounding.\end{IEEEbiography}

\begin{IEEEbiography}[{\includegraphics[width=1in,height=1.25in,clip,keepaspectratio]{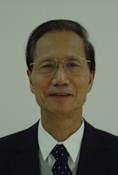}}]{Bo Zhang} graduated from the Department of Automatic Control, Tsinghua University, Beijing,
China, in 1958. Currently, he is a Professor in the Department of Computer Science and Technology, Tsinghua University and a Fellow of Chinese Academy of Sciences, Beijing, China. His main interests are artificial intelligence, pattern recognition, neural networks, and intelligent control. He has published over 150 papers and four monographs in these fields.
\end{IEEEbiography}

\begin{IEEEbiography}[{\includegraphics[width=1in,height=1.25in,clip,keepaspectratio]{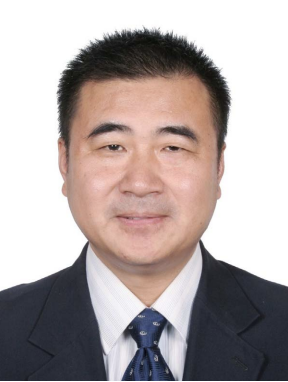}}]{Fuchun Sun} (Fellow, IEEE) received the B.S. and M.S. degrees from the Naval Aeronautical Engineering Academy, Yantai, China, in 1986 and 1989, respectively, and the Ph.D. degree from Tsinghua University, Beijing, China, in 1998. From 1998 to 2000, he was a Post-Doctoral Fellow with the Department of Automation, Tsinghua University, where he is currently a Professor with the
Department of Computer Science and Technology. His current research interests include intelligent control, neural networks, fuzzy systems, variable structure control, nonlinear systems, and robotics.
\end{IEEEbiography}


%
\begin{IEEEbiography}[{\includegraphics[width=1in,height=1.25in,clip,keepaspectratio]{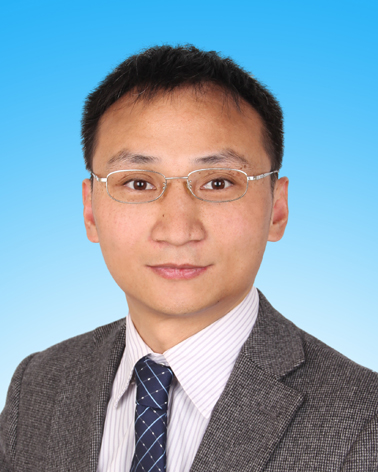}}]{Xiaolin Hu}  (S'01, M'08, SM'13) received the B.E. and M.E. degrees in Automotive Engineering from Wuhan University of Technology, Wuhan, China, and the Ph.D. degree in Automation and Computer-Aided Engineering from The Chinese University of Hong Kong, Hong Kong, China, in 2001, 2004, 2007, respectively. He is now an Associate Professor at the Department of Computer Science and Technology, Tsinghua University, Beijing, China. His current research interests include deep learning and computational neuroscience. Now he is an Associate Editor of IEEE Transactions on Pattern Analysis and Machine Intelligence, IEEE Transactions on Image Processing and Cognitive Neurodynamics.
\end{IEEEbiography}

\setcounter{table}{0}
\renewcommand{\thetable}{A\arabic{table}}
\renewcommand{\theHtable}{Supplement.\thetable}

\setcounter{figure}{0}
\renewcommand{\thefigure}{A\arabic{figure}}
\renewcommand{\theHfigure}{Supplement.\thefigure}

\appendices

\begin{table*}[!htbp]
  \centering
  \caption{Computational cost of ROCK and three ROW models. Computational cost increases slowly with the number of object categories.}
  \small
   \setlength{\tabcolsep}{2pt} {
    \begin{tabular}{c|ccc|ccc}
    \toprule
    \multirow{2}*{\textbf{Target}}   & \multicolumn{3}{c|}{\textbf{Flops}} & \multicolumn{3}{c}{\textbf{Parameters}}\\
    \cmidrule{2-7}
       & \textbf{Face-Body} & \textbf{Pascal-Part-C}  & \textbf{PartImageNet-C} & \textbf{Face-Body} & \textbf{Pascal-Part-C}  & \textbf{PartImageNet-C}\\
    \midrule
    ROW-50     & $4.11G$ & $4.11G$ & $4.11G$ & $23.51M$ & $23.53M$ & $23.76M$ \\
    ROW$\rm{_M}$-50      &  $8.67G$ & $8.67G$ & $8.67G$ & $10.17M$ & $10.18M$ & $10.36M$ \\
    ROW$\rm{_X}$-101      & $8.01G$ & $8.01G$ & $8.01G$ & $42.13M$ & $42.14M$ & $42.38M$ \\ 
    \midrule
    ROCK      & $8.69G$ & $8.71G$ & $9.27G$ & $10.19M$ & $10.22M$ & $10.94M$\\
    \bottomrule
    \end{tabular}
    }

\label{tab:flops}%
\end{table*}%

\begin{table*}[!hbtp]
  \centering
  \caption{\xiaoli{The accuracies (\%) of ROCK combined with AT under five adaptive attacks. All attacks iterated 40 steps by using PGD with step size $\epsilon / 8$. The adversarial examples used by AT were generated by one of the adaptive attacks, \emph{random}, with the same setting as in evaluation.  We highlight the \textbf{lowest} accuracies of ROCK under the five adaptive attacks in each setting.}}
  \label{tab:AttackROCKAT}%
  \small
   \setlength{\tabcolsep}{2.5pt} {
    \begin{tabular}{c||c|cc|c|cc||c|cc|c|cc}
    \toprule
    \textbf{Dataset}   & \multicolumn{12}{c}{\textbf{Pascal-Part-C}}\\
    \midrule
    \textbf{AT Method}                & \multicolumn{3}{c|}{\textbf{PGD-AT}} & \multicolumn{3}{c||}{\textbf{TRADES}}  & \multicolumn{3}{c|}{\textbf{TRADES-AWP}} & \multicolumn{3}{c}{\textbf{TRADES-AWP-EMA}}\\
    \cmidrule{1-13}
     \textbf{Attack} & \textbf{Benign} & \textbf{$\bm{\epsilon =4}$} & \textbf{$\bm{\epsilon =8}$} & \textbf{Benign} & \textbf{$\bm{\epsilon =4}$} & \textbf{$\bm{\epsilon =8}$} & \textbf{Benign} & \textbf{$\bm{\epsilon =4}$} & \textbf{$\bm{\epsilon =8}$} & \textbf{Benign} & \textbf{$\bm{\epsilon =4}$} & \textbf{$\bm{\epsilon =8}$} \\
    \midrule
    \textbf{Random} & \multirow{5}*{$71.5$}& $54.2$& $40.1$ & \multirow{5}*{$70.4$}& $56.0$& $41.9$ & \multirow{5}*{$73.0$}& $58.6$& $43.0$ & \multirow{5}*{$73.4$}& $57.0$& $42.7$ \\
    \textbf{Targeted} & & $57.2$ & $42.4$ & & $58.8$ & $43.7$ & & $61.3$ & $48.5$ & & $60.4$ & $46.8$  \\
    \textbf{Untargeted} & &$56.2$ &$41.9$ & &$58.2$ &$43.2$  & &$61.1$ &$48.2$ & & $60.2$ &$47.6$ \\
    \textbf{Background} & &$56.6$ &$42.5$ & &$57.0$ &$43.5$ & &$59.8$ &$46.0$ & &  $58.3$ &$45.5$  \\
   \textbf{Importance} & & $\bm{50.0}$ &$\bm{29.0}$ & & $\bm{50.9}$ &$\bm{30.8}$ & & $\bm{53.8}$ &$\bm{32.6}$ & & $\bm{53.4}$ & $\bm{32.6}$ \\

   \midrule
   \midrule
    \textbf{Dataset}   & \multicolumn{12}{c}{\textbf{PartImageNet-C}} \\
    \midrule
    \textbf{AT Method}                & \multicolumn{3}{c|}{\textbf{PGD-AT}} & \multicolumn{3}{c||}{\textbf{TRADES}}  & \multicolumn{3}{c|}{\textbf{TRADES-AWP}} & \multicolumn{3}{c}{\textbf{TRADES-AWP-EMA}}\\
    \cmidrule{1-13}
     \textbf{Attack} & \textbf{Benign} & \textbf{$\bm{\epsilon =4}$} & \textbf{$\bm{\epsilon =8}$} & \textbf{Benign} & \textbf{$\bm{\epsilon =4}$} & \textbf{$\bm{\epsilon =8}$} & \textbf{Benign} & \textbf{$\bm{\epsilon =4}$} & \textbf{$\bm{\epsilon =8}$} & \textbf{Benign} & \textbf{$\bm{\epsilon =4}$} & \textbf{$\bm{\epsilon =8}$} \\
    \midrule
    \textbf{Random} & \multirow{5}*{$51.6$}& $37.1$& $20.0$ & \multirow{5}*{$52.8$}& $37.4$& $22.4$ & \multirow{5}*{$54.9$}& $40.1$& $24.2$ & \multirow{5}*{$54.5$}& $39.4$& $24.2$ \\
    \textbf{Targeted} & & $40.1$ &$23.2$ & & $40.9$ &$27.0$ & & $42.6$ & $29.0$ & & $42.7$ &$29.0$ \\
    \textbf{Untargeted} & & $40.7$ &$23.7$ & & $39.3$ &$24.8$& &$41.9$ &$27.3$ & & $41.7$ &$26.8$\\
    \textbf{Background} & &  $37.1$ &$20.2$  & &  $38.4$ &$23.2$ & &$41.0$ &$26.3$ & &  $40.9$ &$26.3$\\
   \textbf{Importance} & & $\bm{31.1}$ & $\bm{14.6}$  & & $\bm{34.1}$ & $\bm{19.3}$ & & $\bm{35.7}$ &$\bm{18.6}$ & & $\bm{35.6}$ & $\bm{18.5}$ \\

   \bottomrule
    \end{tabular}
     }
\end{table*}%

\begin{figure*}[!htbp]
	\centering
 	\includegraphics[width=\linewidth]{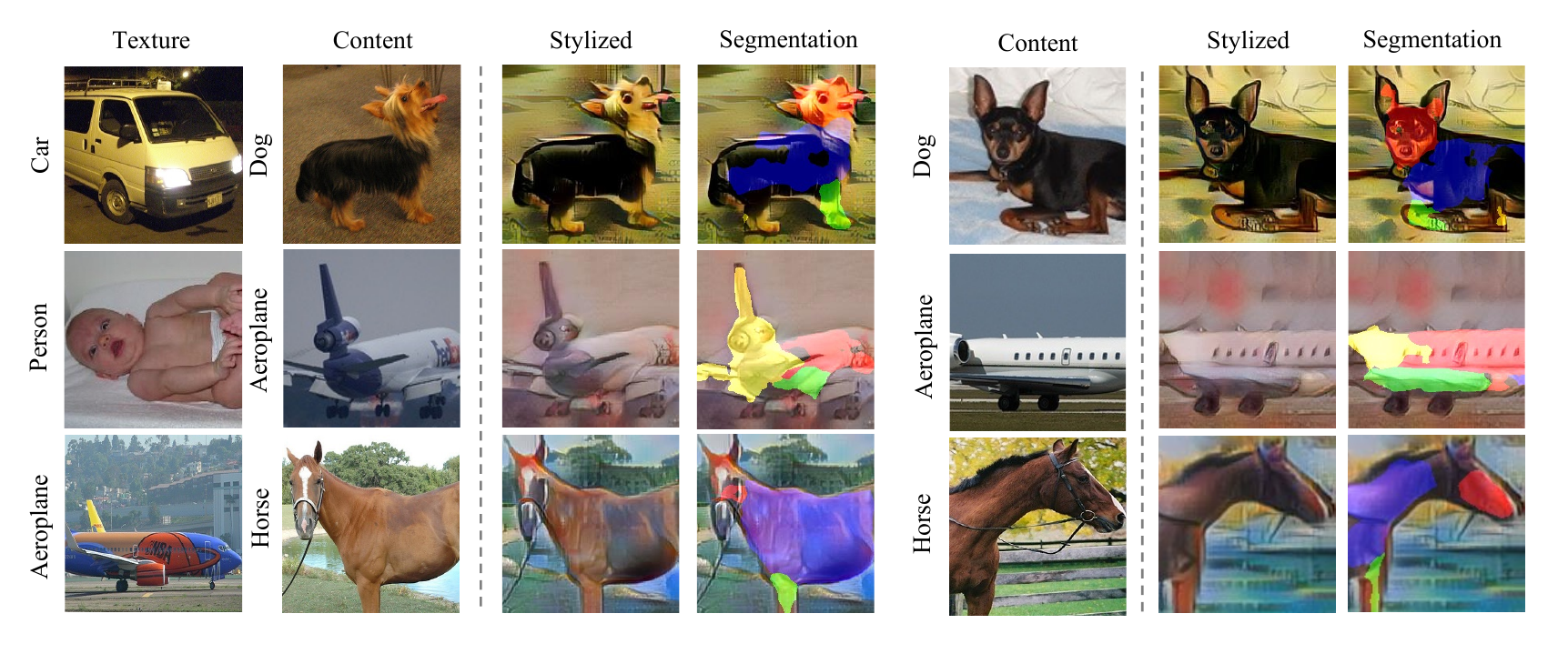}
	\caption{Three other cue conflict images (the third and the sixth column) and their segmentation results by ROCK (the fourth and the seventh column). The columns of \textit{texture} and \textit{content} show the images used to create the cue conflict images. In the first row, we stylize \textit{dog} with \textit{bus}’s texture. In the second row, we stylize \textit{aeroplane} with \textit{human} texture. In the third row, we stylize \textit{horse} with \textit{aeroplane}'s texture.}
	\label{fig:appendix_style}
\end{figure*}

\clearpage
\clearpage

\section{Additional Dataset Processing Details}
\label{sec:apd_dataset}

\subsection{Face-Body} 
CelebAMask-HQ is a human face dataset for face parsing, containing 30,000 face images with part labels. To simplify the task, we removed some small part labels (e.g., eyebrow, denoted by \texttt{Eye\_g}) and merged part labels not related to predefined linkage rules (e.g., \texttt{L\_lip} to \texttt{Mouth}) into eight kinds of different part labels (i.e., \texttt{Nose},\ \texttt{Neck},\ \texttt{L\_eye},\ \texttt{Mouth},\ \texttt{Ear},\ \texttt{Hair},\ \texttt{E\_eye},\ \texttt{Skin}). The mappings between part labels of CelebAMask-HQ and Face-Body are shown in \cref{tab:TypeMapping} (left). LIP is a dataset that focuses on the semantic understanding of persons, containing 24,217 images. We removed images with small resolutions (the width or height less than $100$ pixels). We merged all part labels into six kinds of part labels (i.e., \texttt{head},\ \texttt{cloth},\ \texttt{left-arm},\ \texttt{right-arm},\ \texttt{left-leg},\ \texttt{right-leg}). The mappings are shown in \cref{tab:TypeMapping} (right). As the images of humans contain human faces, we only used the body parts of LIP. That is, we manually cut out images and kept the human body parts in LIP only. Finally, we got 11,767 \textit{body} images. To maintain data balance, we randomly selected 11,767 \textit{face} images from the 30,000 images in CelebAMask-HQ. We divided Face-Body into \emph{train}/\emph{val} with a ratio of 4:1.


\subsection{Pascal-Part-C} 
Since the number of part labels for each object category in Pascal-Part is unbalanced, and some categories (e.g., \emph{chair}) are not even annotated, so we removed categories with labels fewer than 400 to avoid potential risk of underfitting.  As mentioned in the paper, we only used images and labels from 10 categories. Still, some parts of these 10 categories contained many low-quality labels or did not have enough labels, which might harm the performance of segmentation networks. To alleviate this problem, we removed parts with many incorrect labels and merged the remaining parts according to their semantic meanings. The mappings are shown in \cref{tab:TypeMapping4pascal} at the end of the Appendix as it is quite large. After the merging process, one category had about four kinds of part labels on average. Since Pascal-Part contains multiple categories per image, we extracted images according to object bounding boxes to ensure that one image only contained one category. We enlarged each side of the ground-truth bounding box by 15\% in all directions to keep enough background and then used these new bounding boxes to extract the original images. After that, we discarded the extracted images with sides fewer than 40 pixels. Note that the extracted images may still contain multiple categories because of the intersection of objects. To tackle this problem, we discarded the extracted images with an Intersection over Union (IoU) larger than 25\% with other extracted images in the original images. Since the extracted images had many more \textit{person} objects than other objects, we set the IoU threshold to 5\% if the extracted image contained a \textit{person}. Finally, we set all intersection parts of the extracted images to be black. As the size of Pascal-Part-C is relatively small, we divided Pascal-Part-C into \emph{train}/\emph{val} with a ratio of 9:1. \cref{fig:examplemerge}(b) (middle) shows the examples of two categories and \cref{fig:appendix_sample} shows the examples of the other eight categories of Pascal-Part-C and corresponding linkage rules of each category.


\subsection{PartImageNet-C}
\begin{figure*}[!hbt]
	\centering
 	\includegraphics[width=\linewidth]{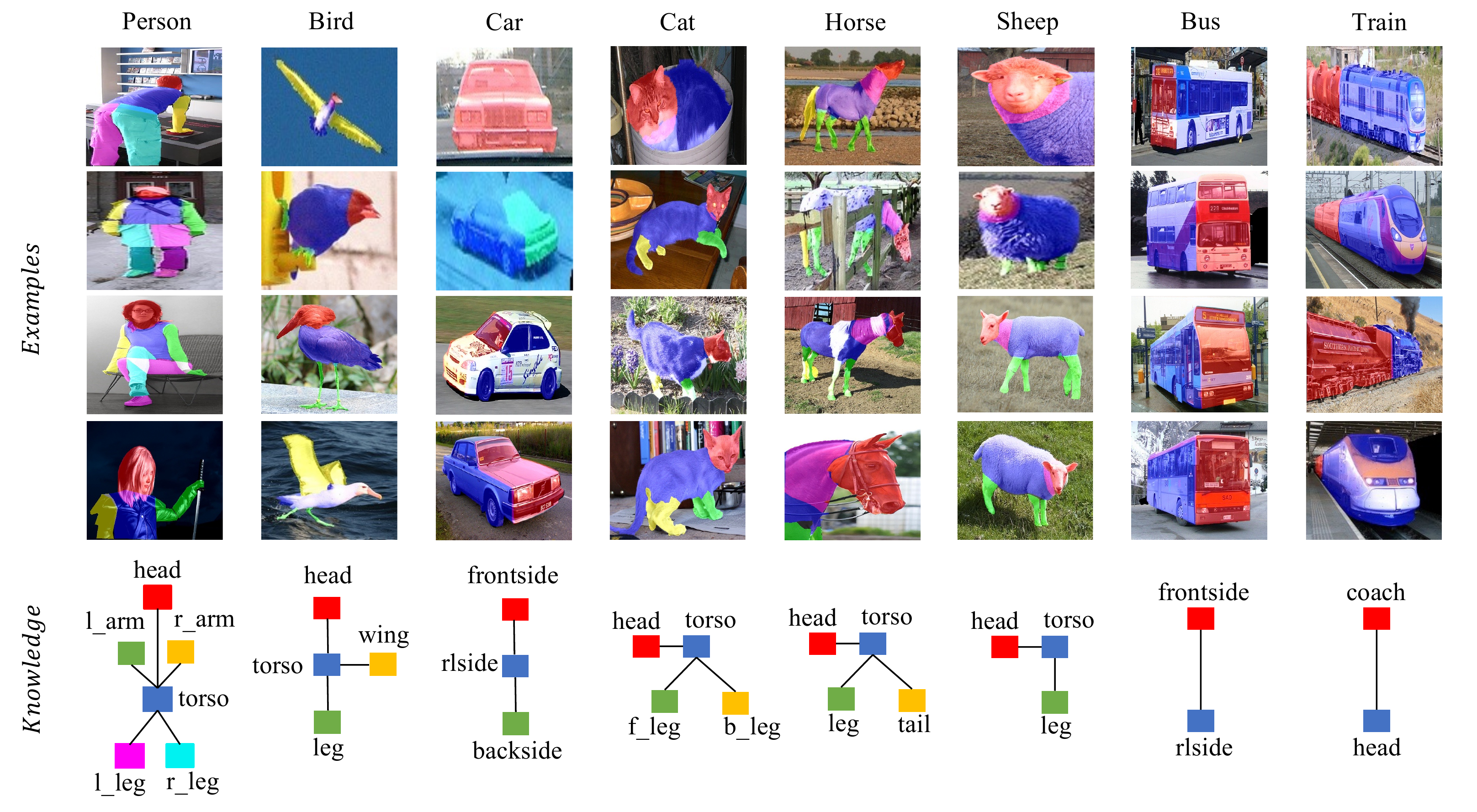}
	\caption{Examples of other eight categories from Pascal-Part-C. In the bottom panels, colored blocks denote object parts, and lines between them denote predefined linkage rules.}
	\label{fig:appendix_sample}
\end{figure*}

PartImageNet includes 158 categories from 11 super-categories and each super-category has similar part labels, e.g, 23 different categories of \emph{cars} are all annotated with \emph{body}, \emph{tier},  and \emph{side mirror}. As mentioned in the paper, we removed certain super-categories with labels unsuitable for part segmentation. For example, \emph{car} is annotated with \emph{side mirror}, \emph{body}, and \emph{tier} only. But \emph{side mirror} is unsuitable for segmentation as it is quite tiny in a image. Pascal-Part-C might give more appropriate labels for a \emph{car} (the second column in \cref{fig:appendix_sample}). After that, 125 categories from eight super-categories were kept. The complete list of categories we used in PartImageNet-C is shown in \cref{tab:PartImageNet-CLabel}. The linkage rules of PartImageNet-C are defined on super-categories due to the characteristics of the part labels (\cref{tab:PartImageNet-CLinkage}) but ROCK distinguished the part labels of several different categories from the same super-category in practice. We divided PartImageNet-C into \emph{train}/\emph{val} with a ratio of 9:1.

\begin{table}[!htp]
\caption{Mappings between part labels in CelebAMask-HQ and Face-Body (left) and mappings between part labels in LIP and Face-Body (right).}
\begin{minipage}[b]{.48\linewidth}
\centering
 \scalebox{0.8}{
\begin{tabular}{c|c}
\toprule
Celebahqmask Parts                   & Face-Body Parts                                                                                                                                                                                                                            \\ \midrule
\texttt{Background}             & \texttt{Background}                                                                                                                                                                                                                        \\ \hline
\texttt{Nose}                & \texttt{Nose}                                                                                                                                                                                                                               \\ \hline
\texttt{Eye\_g}          & \texttt{Background}                                                                                                                                                                                                                                     \\ \hline
\texttt{Cloth}             & \texttt{Background}                                                                                                                                                                                                                              \\ \hline
\texttt{Neck}        & \texttt{Neck}                                                                                                                                                                                                                             \\ \hline
\texttt{L\_eye}      & \texttt{L\_eye}                                                                                                                                                                                                \\ \hline
\texttt{Ear\_r}            & \texttt{Background}                                                                                                                                                                                                     \\ \hline
\texttt{Neck\_l}        & \texttt{Background}                                                                                                                                                                                                             \\ \hline
\texttt{Hat}          & \texttt{Background}                                                                                                                                                                                                                         \\ \hline
\texttt{L\_lip}      & \texttt{Mouth}                                                                                                                                                                                                                          \\ \hline
\texttt{L\_ear}      & \texttt{Ear}                                                                                                                                                                                                                          \\ \hline
\texttt{Mouth}      & \texttt{Mouth}                                                                                                                                                                                                                          \\ \hline
\texttt{R\_brow}      & \texttt{Background}                                                                                                                                                                                                                          \\ \hline
\texttt{U\_lip}      & \texttt{Mouth}                                                                                                                                                                                                                          \\ \hline
\texttt{Hair}      & \texttt{Hair}                                                                                                                                                                                                                          \\ \hline
\texttt{R\_eye}      & \texttt{R\_eye}                                                                                                                                                                                                                          \\ \hline
\texttt{Skin}      & \texttt{Skin}                                                                                                                                                                                                                          \\ \hline
\texttt{L\_brow}      & \texttt{Background}                                                                                                                                                                                                                          \\ \hline
\texttt{R\_ear}      & \texttt{Ear}   
\\ \bottomrule
\end{tabular}
 }
\end{minipage} \hfill
\begin{minipage}[b]{.48\linewidth}
\centering
 \scalebox{0.76}{
\begin{tabular}{c|c}
\toprule
LIP Parts                   & Face-Body Parts                                                                                                                                                                                                                            \\ \midrule
\texttt{Background}             & \texttt{Background}                                                                                                                                                                                                                        \\ \hline
\texttt{Hat}                & \texttt{Head}                                                                                                                                                                                                                               \\ \hline
\texttt{Hair}          & \texttt{Head}                                                                                                                                                                                                                                     \\ \hline
\texttt{Glove}             & \texttt{Background}                                                                                                                                                                                                                              \\ \hline
\texttt{Sunglasses}        & \texttt{Head}                                                                                                                                                                                                                             \\ \hline
\texttt{UpperClothes}      & \texttt{Cloth}                                                                                                                                                                                                \\ \hline
\texttt{Dress}            & \texttt{Cloth}                                                                                                                                                                                                     \\ \hline
\texttt{Coat}        & \texttt{Cloth}                                                                                                                                                                                                             \\ \hline
\texttt{Socks}          & \texttt{Background}                                                                                                                                                                                                                         \\ \hline
\texttt{Pants}      & \texttt{Cloth}                                                                                                                                                                                                                          \\ \hline
\texttt{Jumpsuits}      & \texttt{Cloth}                                                                                                                                                                                                                          \\ \hline
\texttt{Scarf}      & \texttt{Cloth}                                                                                                                                                                                                                          \\ \hline
\texttt{Skirt}      & \texttt{Cloth}                                                                                                                                                                                                                          \\ \hline
\texttt{Face}      & \texttt{Head}                                                                                                                                                                                                                          \\ \hline
\texttt{Left-arm}      & \texttt{Left-arm}                                                                                                                                                                                                                          \\ \hline
\texttt{Right-arm}      & \texttt{Right-arm}                                                                                                                                                                                                                          \\ \hline
\texttt{Left-leg}      & \texttt{Left-leg}                                                                                                                                                                                                                          \\ \hline
\texttt{Right-leg}      & \texttt{Right-leg}                                                                                                                                                                                                                          \\ \hline
\texttt{Left-shoe}      & \texttt{Background}                                                                                                                                                                                                                          \\ \hline
\texttt{Right-shoe}      & \texttt{Background}                                                                                                                                                                                                                          \\ \bottomrule
\end{tabular}
 }
\end{minipage}
\label{tab:TypeMapping}
\end{table}

\begin{table*}[!t]
  \caption{The complete list of categories we used in PartImageNet-C.}
    \centering
    \small
    \begin{tabular}{cccccccc}
    \toprule
     n01693334 & n02002724 & n02492035 & n01632458 & n02492660 & n02097474 & n01630670 & n01440764 \\
     n02006656 & n02397096 & n01855672 & n01739381 & n01688243 & n02096177 & n02109961 & n01695060 \\
     n02444819 & n04482393 & n01729977 & n02130308 & n02441942 & n02071294 & n02102973 & n02489166 \\
     n02483362 & n01828970 & n02058221 & n01742172 & n02422699 & n02423022 & n02124075 & n02129604 \\
     n02486261 & n01694178 & n03792782 & n01687978 & n02101388 & n02033041 & n02488702 & n01608432 \\
     n02112137 & n01664065 & n02090379 & n02536864 & n02486410 & n02091831 & n01665541 & n02025239 \\
     n02133161 & n01749939 & n02484975 & n02422106 & n02356798 & n02412080 & n01692333 & n01669191 \\
     n02009912 & n02132136 & n02443114 & n02487347 & n01753488 & n02510455 & n02481823 & n01728572 \\
     n02690373 & n02102040 & n02098105 & n02415577 & n01744401 & n02089867 & n01641577 & n02480855 \\
     n02096585 & n01644900 & n02607072 & n01740131 & n01729322 & n02100583 & n02128385 & n01443537 \\
     n02493793 & n01756291 & n01735189 & n02114367 & n01698640 & n01755581 & n04509417 & n02099601 \\
     n02442845 & n02009229 & n02125311 & n02483708 & n01748264 & n01685808 & n02514041 & n01614925 \\
     n01689811 & n02109525 & n02490219 & n01484850 & n02134084 & n01843065 & n01824575 & n02101006 \\
     n01667778 & n02092339 & n02134418 & n01667114 & n02085782 & n02120079 & n04552348 & n02017213 \\
     n02493509 & n01491361 & n02447366 & n01494475 & n02655020 & n02494079 & n02417914 & n01728920 \\
     n02480495 & n02835271 & n01697457 & n01644373 & n01734418 &           &           &           \\
    \bottomrule
    \end{tabular}
    \label{tab:PartImageNet-CLabel}
\end{table*}

\begin{table*}[!t]
\caption{The complete list of linkage rules on PartImageNet-C. The linkage rules are defined on super-categories.}
    \centering
    \begin{tabular}{c|c|c|c}
    \toprule
     \textbf{Quadruped} & \textbf{Biped} & \textbf{Fish} & \textbf{Bird} \\
\midrule

 Head $\longleftrightarrow$ Body & Head $\longleftrightarrow$ Body & Head $\longleftrightarrow$ Body & Head $\longleftrightarrow$ Body \\
 Foot $\longleftrightarrow$ Body & Body $\longleftrightarrow$ Hand & Body $\longleftrightarrow$ Fin & Wing $\longleftrightarrow$ Body \\
 Tail $\longleftrightarrow$ Body & Foot $\longleftrightarrow$ Body & Tail $\longleftrightarrow$ Body & Body $\longleftrightarrow$ Tail \\
   & Tail $\longleftrightarrow$ Body &   & Body $\longleftrightarrow$ Foot \\
   \midrule
    \textbf{Snake} & \textbf{Reptile} & \textbf{Bicycle} & \textbf{Aeroplane} \\
\midrule

 Head $\longleftrightarrow$ Body & Head $\longleftrightarrow$ Body & Body $\longleftrightarrow$ Head & Head $\longleftrightarrow$ Body \\
   & Foot $\longleftrightarrow$ Body & Seat $\longleftrightarrow$ Body & Body $\longleftrightarrow$ Engine \\
   & Tail $\longleftrightarrow$ Body & Tier $\longleftrightarrow$ Body & Wing $\longleftrightarrow$ Body \\
   &   &   & Tail $\longleftrightarrow$ Body \\
    \bottomrule
    \end{tabular}
    \label{tab:PartImageNet-CLinkage}
\end{table*}

\begin{table*}[!htbp]
\caption{Mappings between part labels in Pascal-Part and Pascal-Part-C.}
\begin{minipage}[b]{0.323333\textwidth}
\centering
\small
\scalebox{0.8}{
\begin{tabular}{c|c}
\toprule
Pascal-Part Parts & Pascal-Part-C Parts \\ \midrule
\texttt{background} & \texttt{background} \\ \hline
\texttt{car\_fliplate} & \texttt{background} \\ \hline
\texttt{car\_frontside} & \texttt{car\_frontside} \\ \hline
\texttt{car\_rightside} & \texttt{car\_rlside} \\ \hline
\texttt{car\_door} & \texttt{car\_rlside} \\ \hline
\texttt{car\_rightmirror} & \texttt{background} \\ \hline
\texttt{car\_headlight} & \texttt{background} \\ \hline
\texttt{car\_wheel} & \texttt{car\_rlside} \\ \hline
\texttt{car\_window} & \texttt{background} \\ \hline
\texttt{car\_leftmirror} & \texttt{background} \\ \hline
\texttt{car\_backside} & \texttt{car\_backside} \\ \hline
\texttt{car\_leftside} & \texttt{car\_rlside} \\ \hline
\texttt{car\_roofside} & \texttt{background} \\ \hline
\texttt{car\_bliplate} & \texttt{background} \\ \hline
\texttt{bird\_head} & \texttt{bird\_head} \\ \hline
\texttt{bird\_leye} & \texttt{background} \\ \hline
\texttt{bird\_beak} & \texttt{bird\_head} \\ \hline
\texttt{bird\_torso} & \texttt{bird\_torso} \\ \hline
\texttt{bird\_lwing} & \texttt{bird\_wing} \\ \hline
\texttt{bird\_tail} & \texttt{bird\_torso} \\ \hline
\texttt{bird\_reye} & \texttt{background} \\ \hline
\texttt{bird\_neck} & \texttt{bird\_head} \\ \hline
\texttt{bird\_rwing} & \texttt{bird\_wing} \\ \hline
\texttt{bird\_lleg} & \texttt{bird\_leg} \\ \hline
\texttt{bird\_lfoot} & \texttt{bird\_leg} \\ \hline
\texttt{bird\_rleg} & \texttt{bird\_leg} \\ \hline
\texttt{bird\_rfoot} & \texttt{bird\_leg} \\ \hline
\texttt{person\_head} & \texttt{person\_head} \\ \hline
\texttt{person\_torso} & \texttt{person\_torso} \\ \hline
\texttt{person\_ruarm} & \texttt{person\_rarm} \\ \hline
\texttt{person\_rlleg} & \texttt{person\_rleg} \\ \hline
\texttt{person\_ruleg} & \texttt{person\_rleg} \\ \hline
\texttt{person\_rfoot} & \texttt{person\_rleg} \\ \hline
\texttt{person\_leye} & \texttt{background} \\ \hline
\texttt{person\_mouth} & \texttt{background} \\ \hline
\texttt{person\_hair} & \texttt{background} \\ \hline
\texttt{person\_nose} & \texttt{background} \\ \hline
\texttt{person\_llarm} & \texttt{person\_larm} \\ \hline
\texttt{person\_luarm} & \texttt{person\_larm} \\ \hline
\texttt{person\_lhand} & \texttt{person\_larm} \\ \hline
\texttt{person\_neck} & \texttt{person\_head} \\ \hline
\texttt{person\_luleg} & \texttt{person\_lleg} \\ \hline
\texttt{person\_rear} & \texttt{background} \\ \hline
\texttt{person\_reye} & \texttt{background} \\ \hline
\texttt{person\_lebrow} & \texttt{background} \\ \hline
\texttt{person\_rebrow} & \texttt{background} \\ \hline
\texttt{person\_rlarm} & \texttt{person\_rarm} \\ \hline
\texttt{person\_llleg} & \texttt{person\_lleg} \\ \hline
\texttt{person\_lfoot} & \texttt{person\_lleg} \\ \hline
\texttt{person\_lear} & \texttt{background} \\ \hline
\texttt{person\_rhand} & \texttt{person\_rarm} \\ \hline
\texttt{cat\_head} & \texttt{cat\_head} \\ \hline
\texttt{cat\_lear} & \texttt{background} \\ \bottomrule
\end{tabular}
}
\end{minipage} \hfill
\begin{minipage}[b]{0.323333\textwidth}
\centering
\small
\scalebox{0.8}{
\begin{tabular}{c|c}
\toprule
Pascal-Part Parts & Pascal-Part-C Parts \\ \midrule
\texttt{cat\_rear} & \texttt{background} \\ \hline
\texttt{cat\_leye} & \texttt{background} \\ \hline
\texttt{cat\_reye} & \texttt{background} \\ \hline
\texttt{cat\_nose} & \texttt{background} \\ \hline
\texttt{cat\_torso} & \texttt{cat\_torso} \\ \hline
\texttt{cat\_neck} & \texttt{cat\_head} \\ \hline
\texttt{cat\_lbleg} & \texttt{cat\_bleg} \\ \hline
\texttt{cat\_lbpa} & \texttt{cat\_bleg} \\ \hline
\texttt{cat\_rbleg} & \texttt{cat\_bleg} \\ \hline
\texttt{cat\_rbpa} & \texttt{cat\_bleg} \\ \hline
\texttt{cat\_tail} & \texttt{cat\_torso} \\ \hline
\texttt{cat\_lfleg} & \texttt{cat\_fleg} \\ \hline
\texttt{cat\_lfpa} & \texttt{cat\_fleg} \\ \hline
\texttt{cat\_rfleg} & \texttt{cat\_fleg} \\ \hline
\texttt{cat\_rfpa} & \texttt{cat\_fleg} \\ \hline
\texttt{aeroplane\_body} & \texttt{aeroplane\_body} \\ \hline
\texttt{aeroplane\_lwing} & \texttt{aeroplane\_lwing} \\ \hline
\texttt{aeroplane\_engine} & \texttt{aeroplane\_body} \\ \hline
\texttt{aeroplane\_stern} & \texttt{aeroplane\_stern} \\ \hline
\texttt{aeroplane\_wheel} & \texttt{background} \\ \hline
\texttt{aeroplane\_rwing} & \texttt{aeroplane\_rwing} \\ \hline
\texttt{aeroplane\_tail} & \texttt{aeroplane\_stern} \\ \hline
\texttt{sheep\_head} & \texttt{sheep\_head} \\ \hline
\texttt{sheep\_lear} & \texttt{background} \\ \hline
\texttt{sheep\_rear} & \texttt{background} \\ \hline
\texttt{sheep\_leye} & \texttt{background} \\ \hline
\texttt{sheep\_reye} & \texttt{background} \\ \hline
\texttt{sheep\_lhorn} & \texttt{sheep\_head} \\ \hline
\texttt{sheep\_rhorn} & \texttt{sheep\_head} \\ \hline
\texttt{sheep\_muzzle} & \texttt{background} \\ \hline
\texttt{sheep\_torso} & \texttt{sheep\_torso} \\ \hline
\texttt{sheep\_neck} & \texttt{sheep\_head} \\ \hline
\texttt{sheep\_lflleg} & \texttt{sheep\_leg} \\ \hline
\texttt{sheep\_lfuleg} & \texttt{sheep\_leg} \\ \hline
\texttt{sheep\_lblleg} & \texttt{sheep\_leg} \\ \hline
\texttt{sheep\_lbuleg} & \texttt{sheep\_leg} \\ \hline
\texttt{sheep\_rblleg} & \texttt{sheep\_leg} \\ \hline
\texttt{sheep\_rbuleg} & \texttt{sheep\_leg} \\ \hline
\texttt{sheep\_rflleg} & \texttt{sheep\_leg} \\ \hline
\texttt{sheep\_rfuleg} & \texttt{sheep\_leg} \\ \hline
\texttt{sheep\_tail} & \texttt{sheep\_torso} \\ \hline
\texttt{dog\_head} & \texttt{dog\_head} \\ \hline
\texttt{dog\_lear} & \texttt{background} \\ \hline
\texttt{dog\_rear} & \texttt{background} \\ \hline
\texttt{dog\_leye} & \texttt{background} \\ \hline
\texttt{dog\_reye} & \texttt{background} \\ \hline
\texttt{dog\_muzzle} & \texttt{background} \\ \hline
\texttt{dog\_nose} & \texttt{background} \\ \hline
\texttt{dog\_torso} & \texttt{dog\_torso} \\ \hline
\texttt{dog\_lfleg} & \texttt{dog\_fleg} \\ \hline
\texttt{dog\_lfpa} & \texttt{dog\_fleg} \\ \hline
\texttt{dog\_rfleg} & \texttt{dog\_fleg} \\ \hline
\texttt{dog\_rfpa} & \texttt{dog\_fleg} \\ \bottomrule
\end{tabular}
}
\end{minipage} \hfill
\begin{minipage}[b]{0.323333\textwidth}
\centering
\small
\scalebox{0.8}{
\begin{tabular}{c|c}
\toprule
Pascal-Part Parts & Pascal-Part-C Parts \\ \midrule
\texttt{dog\_lbleg} & \texttt{dog\_bleg} \\ \hline
\texttt{dog\_lbpa} & \texttt{dog\_bleg} \\ \hline
\texttt{dog\_tail} & \texttt{dog\_torso} \\ \hline
\texttt{dog\_rbleg} & \texttt{dog\_bleg} \\ \hline
\texttt{dog\_neck} & \texttt{dog\_head} \\ \hline
\texttt{dog\_rbpa} & \texttt{dog\_bleg} \\ \hline
\texttt{train\_coach} & \texttt{train\_coach} \\ \hline
\texttt{train\_head} & \texttt{train\_head} \\ \hline
\texttt{train\_headlight} & \texttt{train\_head} \\ \hline
\texttt{train\_hfrontside} & \texttt{train\_head} \\ \hline
\texttt{train\_hleftside} & \texttt{train\_head} \\ \hline
\texttt{train\_cleftside} & \texttt{train\_coach} \\ \hline
\texttt{train\_hroofside} & \texttt{train\_head} \\ \hline
\texttt{train\_cfrontside} & \texttt{train\_coach} \\ \hline
\texttt{train\_crightside} & \texttt{train\_coach} \\ \hline
\texttt{train\_hrightside} & \texttt{train\_head} \\ \hline
\texttt{train\_croofside} & \texttt{train\_coach} \\ \hline
\texttt{train\_cbackside} & \texttt{train\_coach} \\ \hline
\texttt{train\_hbackside} & \texttt{train\_head} \\ \hline
\texttt{bus\_headlight} & \texttt{background} \\ \hline
\texttt{bus\_roofside} & \texttt{background} \\ \hline
\texttt{bus\_fliplate} & \texttt{bus\_frontside} \\ \hline
\texttt{bus\_leftside} & \texttt{bus\_lrside} \\ \hline
\texttt{bus\_bliplate} & \texttt{background} \\ \hline
\texttt{bus\_backside} & \texttt{bus\_frontside} \\ \hline
\texttt{bus\_frontside} & \texttt{bus\_frontside} \\ \hline
\texttt{bus\_rightside} & \texttt{bus\_lrside} \\ \hline
\texttt{bus\_door} & \texttt{bus\_lrside} \\ \hline
\texttt{bus\_leftmirror} & \texttt{background} \\ \hline
\texttt{bus\_rightmirror} & \texttt{background} \\ \hline
\texttt{bus\_wheel} & \texttt{background} \\ \hline
\texttt{bus\_window} & \texttt{background} \\ \hline
\texttt{horse\_head} & \texttt{horse\_head} \\ \hline
\texttt{horse\_lear} & \texttt{background} \\ \hline
\texttt{horse\_rear} & \texttt{background} \\ \hline
\texttt{horse\_leye} & \texttt{background} \\ \hline
\texttt{horse\_muzzle} & \texttt{background} \\ \hline
\texttt{horse\_rbho} & \texttt{horse\_leg} \\ \hline
\texttt{horse\_torso} & \texttt{horse\_torso} \\ \hline
\texttt{horse\_neck} & \texttt{horse\_head} \\ \hline
\texttt{horse\_lfuleg} & \texttt{horse\_leg} \\ \hline
\texttt{horse\_rfuleg} & \texttt{horse\_leg} \\ \hline
\texttt{horse\_lblleg} & \texttt{horse\_leg} \\ \hline
\texttt{horse\_lbuleg} & \texttt{horse\_leg} \\ \hline
\texttt{horse\_lbho} & \texttt{horse\_leg} \\ \hline
\texttt{horse\_rblleg} & \texttt{horse\_leg} \\ \hline
\texttt{horse\_rbuleg} & \texttt{horse\_leg} \\ \hline
\texttt{horse\_reye} & \texttt{background} \\ \hline
\texttt{horse\_lflleg} & \texttt{horse\_leg} \\ \hline
\texttt{horse\_lfho} & \texttt{horse\_leg} \\ \hline
\texttt{horse\_rflleg} & \texttt{horse\_leg} \\ \hline
\texttt{horse\_rfho} & \texttt{horse\_leg} \\ \hline
\texttt{horse\_tail} & \texttt{horse\_tail} \\ \bottomrule
\end{tabular}
}
\end{minipage} \hfill
\label{tab:TypeMapping4pascal}

\end{table*}

\end{document}